\documentclass{article}
\usepackage[english]{babel}
\usepackage[utf8]{inputenc}

\usepackage[letterpaper,top=2cm,bottom=2cm,left=3cm,right=3cm,marginparwidth=1.75cm]{geometry}

\usepackage{amsmath}
\usepackage{graphicx}
\usepackage[colorlinks=true, allcolors=blue]{hyperref}
\usepackage{xcolor}
\usepackage{lineno}
\usepackage{authblk}
\usepackage{pifont}
\usepackage{soul}
\usepackage{rotating}
\usepackage{comment}
\usepackage[section]{placeins}
\usepackage{hyperref}
\usepackage{setspace}
\usepackage{caption}
\usepackage{amssymb}
\usepackage{float}

\usepackage[
  backend=biber,
  bibstyle=ieee,
  citestyle=ieee-comp,
  sorting=none,
  maxbibnames=5,
  minbibnames=1
]{biblatex}
\usepackage{csquotes}
\usepackage[none]{hyphenat}
\title{
Personalized Emotional Intelligence in Generative AI through Symbolic Affective Reasoning
}

\author {
    Qing Lin, \hspace{5mm}
    Mengmi Zhang
    \vspace{2mm}\\
    \small College of Computing and Data Science, Nanyang Technological University, Singapore,\\
    \small Address correspondence to: mengmi.zhang@ntu.edu.sg\\
}
\date{}

\begin{document}
\maketitle

\textbf{Text statistics}

6 figures

9 supplementary figures

\begin{abstract}
Emotional intelligence enables humans to recognize emotions, infer their causes, reason about interventions, and modify their environment to achieve desired affective states. Despite recent advances in artificial intelligence (AI), current models remain largely limited to generating realistic content or performing semantic reasoning, with little capacity for understanding, predicting, and personalizing human emotional responses. Here we introduce Emotion-augmented geneRatiOn System (EROS), a hybrid AI framework that integrates symbolic reasoning with deep learning to enable personalized emotion augmentation through visual content. Leveraging large-scale image–emotion datasets, EROS discovers generalizable affective rules, identifies emotion-relevant image regions, and predicts context-aware visual modifications that preserve scene semantics while steering emotional responses toward desired targets. To account for individual variability, EROS incorporates an expandable memory bank that supports inference-time personalization without model fine-tuning, yielding interpretable emotional profiles and rapid adaptation to new users. Across extensive human psychophysics experiments, EROS elicits target emotional responses more effectively than state-of-the-art large multimodal models while adapting to individual affective preferences. Beyond affective computing, EROS provides a foundation for AI systems that can understand, reason about, and augment human cognitive states, with potential applications in mental health, adaptive media, education, and human–computer interaction.

\end{abstract}

\newpage

\setstretch{1.5}
\section{Introduction}

When communicating with a homesick friend, one may selectively emphasize warm lighting, familiar objects, or comforting visual cues in a photograph to evoke a sense of belonging (\textbf{Fig.\ref{fig:fig1}A}). This ability to infer another person's emotional state, understand how visual elements shape affect, reason about potential interventions, and modify visual content to achieve a desired emotional outcome is a hallmark of human emotional intelligence \cite{mayer2004target, mayer2008human, mayer2001emotional, salovey2008positive} (\textbf{Fig.\ref{fig:fig1}B}).

Computationally instantiating such emotional intelligence in machines remains a fundamental challenge. Emotional responses do not arise from individual objects in isolation but emerge from the holistic interpretation of scenes and their contextual relationships. In the example above (\textbf{Fig.\ref{fig:fig1}A}), lighting, familiar objects, spatial layout, and implied context jointly contribute to the perceived emotion, requiring models to reason about interactions among visual elements and their meanings \cite{gerdes2014emotional, travere2023interplay, barsalou1999perceptions, cross2016shaping, nook2015new, yarosh2022perceptions, yang2023emoset}. Moreover, emotional perception is inherently personalized. The same image may evoke comfort and warmth for one observer but loneliness, fear, or indifference for another, reflecting differences in personal experience, cultural background, and affective traits \cite{barrett2019emotions, ellsworth2003appraisal, clore2013psychological}. Finally, emotional regulation through image modification is fundamentally underconstrained: many possible edits can induce similar emotional outcomes, yet only a subset preserves the semantic content and structural integrity of the original scene \cite{barrett2011context, chen2022inferential, zhou2025object}.

Recent advances in AI have enabled models to generate photorealistic images \cite{ktena2024generative, siddals2024happened, swanson2024generative}, learn increasingly powerful visual representations for recognition and localization \cite{tee2023integrating, talbot2023tuned, shi2025unveiling, singh2023learning, krizhevsky2012imagenet, simonyan2014very, he2016resnet, dosovitskiy2020image, ding2022efficient, wang2023object, han2024flow, cai2025learning, wu2023label, wang2024pose, wang2025gazing}, and
support scene understanding and structured visual reasoning \cite{lu2024multimodal, jia2025seeing, khandelwal2023adaptive, zhang2020putting, bomatter2021pigs, liu2022reason}.
Despite these achievements, contemporary AI systems remain largely optimized for semantic understanding rather than affective understanding. They can recognize what is present in an image, but have limited ability to explain why visual content evokes specific emotions, predict how emotional responses vary across individuals, or generate context-aware modifications that reliably achieve desired affective outcomes.

To bridge this gap, we introduce the problem of \textit{personalized affective image editing}, which evaluates an AI system's ability to perceive, understand, reason about, and modify visual content to achieve targeted emotional outcomes. Given a source image and a target emotional valence (positive or negative), the goal is to generate minimally modified images that reliably evoke the desired affect in a specific observer while preserving the semantic meaning and structural consistency of the original scene (\textbf{Fig.\ref{fig:fig1}C}).

To investigate whether emotional intelligence can be computationally instantiated, we propose the \textit{Emotion-augmented geneRatiOn System} (EROS), a hybrid framework that integrates symbolic reasoning with deep learning (\textbf{Fig.\ref{fig:fig2}A}; \textbf{Sec.\ref{sec:EROS}}). Rather than relying solely on end-to-end optimization, EROS derives structured affective knowledge from large-scale image--emotion datasets by identifying recurring patterns that capture how objects, attributes, interactions, and scene context jointly contribute to emotional responses. These patterns are organized into a compositional affective knowledge structure, termed the \textit{EmoTree}, which links interpretable visual motifs to predictable emotional outcomes.

Building upon this knowledge base, EROS acquires individualized emotional profiles through iterative user interaction and feedback. These profiles are learned and stored in an explicit memory bank that supports inference-time personalization without model fine-tuning, enabling rapid adaptation to new users while maintaining interpretability. The resulting framework combines the transparency of symbolic reasoning with the flexibility of data-driven learning, allowing emotion-aware image modification that is both context-sensitive and personalized.

EROS differs fundamentally from existing approaches (\textbf{Sec.\ref{sec:baselines}}). Prior affective image-editing methods primarily rely on global style transfer or prompt-based generation and therefore lack explicit reasoning about the compositional structure of scenes \cite{galanos2021affectgan, pitie2007automated, gatys2015neural, kwon2022clipstyler, weng2023affective, lin2025make, yang2025emoedit, zhang2026emokgedit, zhang2026affective}. 
Symbolic affective systems typically depend on manually curated associations and do not model interactions among visual elements within complex real-world environments \cite{borth2013large, jou2015visual, valitutti2004developing, bowers1993nonverbal, cambria2017affective, you2015robust, buechel2017emobank, mohammad2010emotions, mohammad2013crowdsourcing}. 
Meanwhile, large multimodal models \cite{hurst2024gpt, openai2025o4, GPT-Image1, nanobanana} offer flexible image editing capabilities but often suffer from limited interpretability \cite{elyoseph2024capacity, wang2025llms, sofroniew2026emotion, barez2025chain, lian2024gpt}, unreliable affective reasoning \cite{feuerriegel2023research, xing2025emotionhallucer, bojic2024signs, turpin2023language, huang2025survey, vzorinab2024emotional, bai2024hallucination}, and expensive personalization through extensive fine-tuning \cite{sun2025other, murtaza2022ai, wang2024exploring, abbes2024generative, pataranutaporn2021ai, habicht2024closing, Supermemory}. In contrast, EROS integrates interpretable symbolic representations with scalable learning to enable accurate, context-aware, and personalized affective editing while remaining transparent, data-efficient, and independent of proprietary foundation models.

The ability to understand and augment human emotions has implications far beyond image editing. Emotionally intelligent AI systems could support mental healthcare through emotion-aware interventions, improve human--AI interaction through adaptive and empathetic interfaces, personalize educational content according to learner affect, and facilitate emotionally adaptive media creation and communication. By operating locally and supporting privacy-preserving personalization, EROS provides a practical foundation for these applications.

We evaluate EROS through six human psychophysics experiments comprising 33,380 trials collected from 296 participants (\textbf{Sec.\ref{sec:exp_all}}). Across tasks spanning emotion recognition, affective reasoning, causal intervention, personalization, and image editing, EROS consistently aligns more closely with human judgments than existing approaches. EROS accurately identifies emotion-evoking image regions, generates interpretable and context-preserving modifications, and elicits faster and stronger target emotional responses than state-of-the-art baselines, including large multimodal models. Through iterative interaction, it also acquires stable individualized affective profiles that generalize across semantically related scenes.

Together, these contributions establish a comprehensive framework for studying emotional intelligence in artificial systems. Supported by large-scale human psychophysics data, standardized evaluation protocols, and quantitative benchmarks, our work provides a foundation for developing AI systems that can perceive, understand, reason about, and augment human emotional states. More broadly, it provides a scalable blueprint for developing AI capable of augmenting a broad range of human cognitive and affective functions.

\section{Results} \label{sec:results}

We address the problem of personalized affective image editing (\textbf{Fig.\ref{fig:fig1}A}) through the proposed Emotion-augmented geneRatiOn System (EROS), a hybrid framework that combines symbolic reasoning with deep learning (\textbf{Fig.\ref{fig:fig2}A}; \textbf{Sec.\ref{sec:EROS}}). To evaluate whether emotional intelligence can be computationally instantiated in artificial systems, we systematically assess EROS across its core capabilities (\textbf{Fig.~\ref{fig:fig1}B}), including emotion recognition, affective understanding and reasoning, controllable emotion regulation through affective image editing, and personalization via human feedback. Across six human psychophysics experiments comprising 33,380 trials from 296 participants (\textbf{Sec.\ref{sec:exp_all}}), we benchmark EROS against state-of-the-art affective image editing methods, large multimodal models, and emotionally intelligent baselines (\textbf{Sec.\ref{sec:baselines}}) using standardized experimental protocols (\textbf{Fig.\ref{fig:fig1}C}) and quantitative evaluation metrics (\textbf{Sec.\ref{sec:data_analysis}}).

\subsection{EROS outperforms existing affective image editing methods}

\subsubsection{EROS elicits targeted emotions while preserving scene structure}
\label{sec:pairwise}
\begin{figure}[H]
\begin{center}
\includegraphics[width=0.95\linewidth]{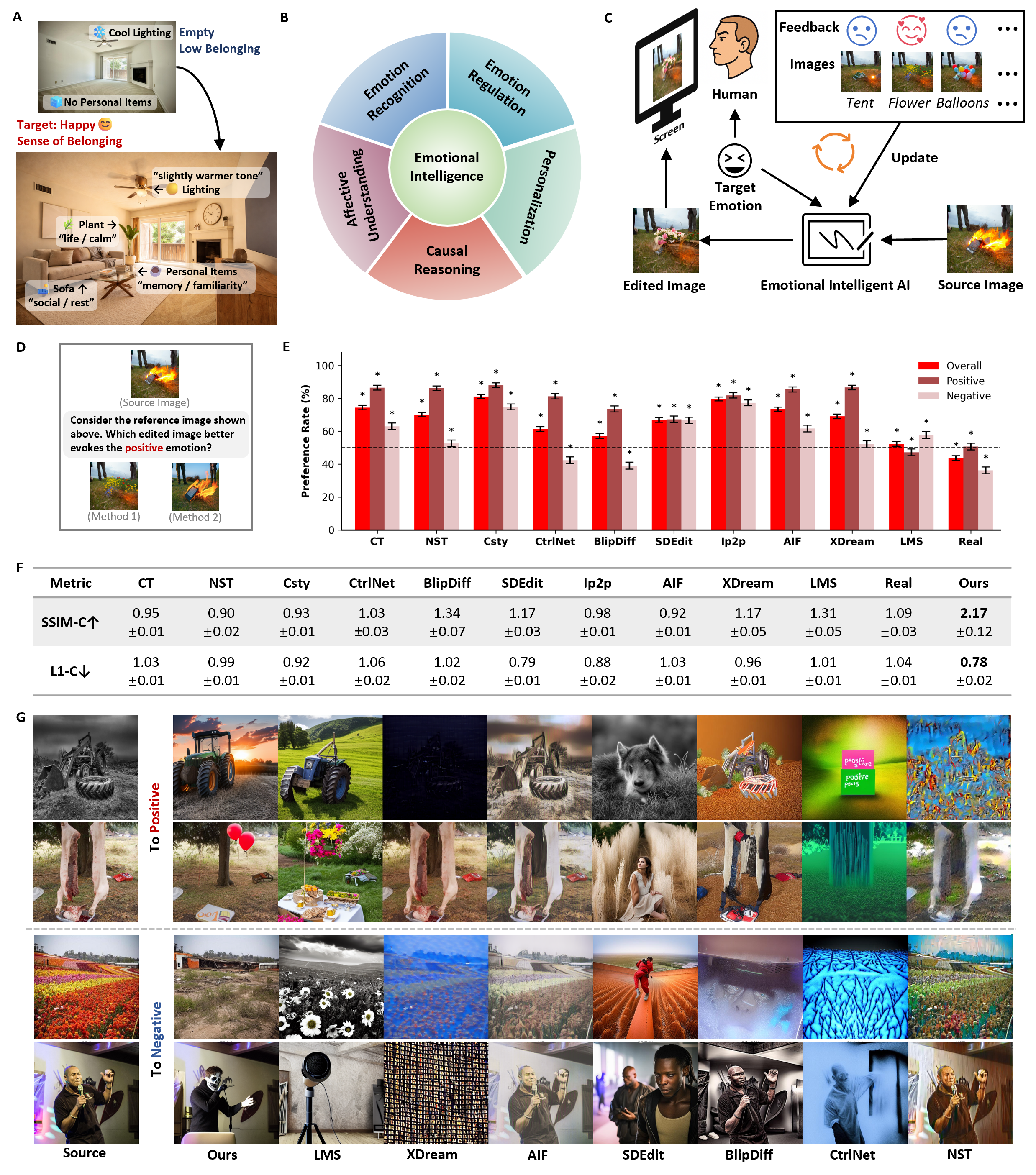}\vspace{-8mm}
\end{center}
   \caption{\footnotesize
    \textbf{Operationalizing emotional intelligence in humans and machines.}
    \textbf{A}, 
    A concrete example illustrating emotional intelligence in humans. Given a source image and a target emotional goal (e.g., happy), humans identify emotion-relevant elements (e.g., cool lighting), reason about the causes of the current emotional response, and selectively modify visual cues (e.g., lighting, objects, and contextual details) to achieve the desired affective outcome while preserving scene coherence.
    \textbf{B}, 
    Conceptual framework of emotional intelligence. We operationalize emotional intelligence as five interconnected capabilities: emotion recognition, affective understanding, causal reasoning, personalization, and emotion regulation.
    \textbf{C}, 
    Human-in-the-loop framework for benchmarking emotionally intelligent AI. Given a source image and target emotion, the AI system generates edited images, receives human feedback, updates its affective knowledge and personalized preferences, and iteratively refines future edits.
    \textbf{D,} Schematic illustration of the Exp-EmoPairwise experiment (\textbf{Sec.\ref{sec:Exp-EmoPairwise}}). Given a source image, a target valence, and two edited images, participants select the image that better evokes the target valence while preserving the semantic content and structural integrity of the source image. In each trial, one image is generated by our proposed Emotion-augmented geneRatiOn System (EROS) and the other by a competing emotionally intelligent AI system. 
    \textbf{E},
    Human preference for images edited by our proposed EROS compared with other emotionally intelligent AI systems in the Exp-EmoPairwise experiment. Bars show preference rates aggregated across all target valences (red), positive-valence trials (dark red), and negative-valence trials (light red). The dashed line indicates chance performance (50\%). Error bars denote standard errors. Statistical significance is indicated by * ($p < 0.05$), whereas non-significant comparisons are denoted by n.s.
    \textbf{F},
    Quantitative comparison of structural fidelity across all emotionally intelligent AI methods. Higher SSIM-C and lower L1-C (\textbf{Sec.\ref{sec:metric}}) indicate that modifications are concentrated within emotion-relevant regions while preserving the semantic content and structural integrity of the remaining image.
    \textbf{G}, Representative affective image-editing results. Four example source images (rows) are edited toward the target valence using EROS (ours) and 7 top emotionally intelligent AI systems (columns). Results from the remaining 3 baseline methods and additional qualitative examples are provided in \textbf{Fig.\ref{fig:S1}}.      
    }
    \vspace{-5mm}
\label{fig:fig1}
\end{figure}

We first evaluate whether EROS generates affective image edits that more effectively evoke the target emotional response than existing approaches (\textbf{Sec.\ref{sec:baselines}}). Because most state-of-the-art methods neither support iterative refinement nor incorporate user personalization, we perform comparisons in a single-step setting using EROS-1loop (\textbf{Sec.\ref{sec:EROS_Variants}}) and the corresponding single-step variants of all baselines (\textbf{Sec.\ref{sec:baselines}}) in the Exp-EmoPairwise experiment (\textbf{Fig.~\ref{fig:fig1}D}; \textbf{Sec.~\ref{sec:Exp-EmoPairwise}}). In each trial, participants compared an image generated by EROS-1loop with one generated by a competing method and selected the edit that better achieved the target valence while preserving the semantic content and structural integrity of the source image. We quantified performance using the human preference rate of EROS over each competing method (\textbf{Sec.~\ref{sec:metric}}).

In parallel, to evaluate whether affective editing preserves the original scene while selectively modifying emotion-relevant content, we measure structural fidelity using two weighted similarity metrics (\textbf{Sec.~\ref{sec:metric}}): SSIM-C ($\uparrow$) and L1-C ($\downarrow$). Unlike conventional image similarity metrics, both explicitly separate edited and preserved regions using human-derived weight maps, which assign higher weights to regions expected to change for inducing the target emotion and lower weights to regions that should remain unchanged. This design enables a more faithful assessment of both structural preservation and emotion-targeted modifications.

As shown in \textbf{Fig.~\ref{fig:fig1}E and F}, EROS achieves the most favorable balance between affective effectiveness and source fidelity, outperforming most existing affective image editing methods in eliciting the target emotional response (\textbf{Fig.~\ref{fig:fig1}E}) while preserving the semantic content and structural integrity of the source image (\textbf{Fig.~\ref{fig:fig1}F}). Human preference rates are significantly above the 50\% chance level for most competing methods, indicating that participants generally preferred images generated by EROS over those produced by state-of-the-art affective editing approaches (Welch's two-tailed t-test, $p < 10^{-3}$). Meanwhile, EROS achieves the highest SSIM-C (2.17) and the lowest L1-C (0.78) among all editing methods, demonstrating that its edits are concentrated within emotion-relevant regions while preserving the remainder of the scene (Welch's two-tailed t-test, $p < 10^{-3}$).

The results reveal complementary limitations of existing affective image editing paradigms. Global transformation methods, including CT\cite{pitie2007automated}, NST\cite{gatys2015neural}, CSty\cite{kwon2022clipstyler}, and AIF\cite{weng2023affective}, primarily manipulate low-level image appearance through color, style, or global filtering, but lack explicit reasoning about the semantic causes of emotion. Consequently, they often fail to induce the desired affective response despite preserving the overall scene. In contrast, prompt-based diffusion methods, including Ip2p\cite{brooks2023instructpix2pix}, SDEdit\cite{meng2021sdedit}, CtrlNet\cite{zhang2023controlnet}, and BlipDiff\cite{li2023blip}, generate semantically richer edits that more effectively alter emotional perception, but frequently modify image regions unrelated to the target emotion, leading to unnecessary structural changes. These contrasting behaviors are quantitatively captured by the proposed SSIM-C and L1-C metrics, which demonstrate that EROS achieves a superior balance between effective emotion augmentation and preservation of the original scene.

Performance further differs across target valence. For positive-valence editing, EROS performs on par with LMS (\textbf{Sec.\ref{sec:baselines}}) and significantly outperforms nearly all other baselines, suggesting that both EROS and large multimodal models can successfully reason about and introduce semantically meaningful positive content. In contrast, EROS substantially outperforms LMS under negative valence. One possible explanation is that proprietary large multimodal models are constrained by safety guardrails that discourage the generation of threatening, violent, or otherwise distressing imagery, thereby limiting their ability to produce strong negative affective transformations. 

In comparison, only NST\cite{gatys2015neural}, CtrlNet\cite{zhang2023controlnet}, and BlipDiff\cite{li2023blip} become more competitive under negative valence. This observation suggests that negative affect can sometimes be induced through visually salient stylistic or structural perturbations without requiring equally strong semantic reasoning. By contrast, EROS induces negative affect through structured, semantically grounded, and localized modifications, enabling competitive emotional effectiveness while maintaining substantially higher structural fidelity.

Finally, we compare EROS with the Real baseline (\textbf{Sec.\ref{sec:baselines}}), which directly retrieves images from EmoSet \cite{yang2023emoset}, a large-scale image--emotion dataset annotated with human valence labels (\textbf{Sec.~\ref{sec:Emoset}}), rather than editing the input image. Real therefore serves as an approximate upper bound for unconstrained emotion elicitation. EROS performs on par with Real under positive valence and only slightly below Real under negative valence, indicating that unrestricted target-valence images can evoke particularly strong emotional responses. However, because Real images bear no semantic or structural correspondence to the source image, their SSIM-C or L1-C are much worse than EROS. In contrast, EROS jointly achieves strong affective alignment and the highest structural fidelity among all editing methods, making it a more practical solution for controllable affective image editing.

Representative examples further illustrate the qualitative differences between EROS and competing affective image editing methods (\textbf{Fig.~\ref{fig:fig1}G}). Owing to space constraints, we show EROS together with the seven strongest competing methods; additional examples and the remaining baselines are provided in \textbf{Fig.~\ref{fig:S1}}.
Across both positive- and negative-valence editing, EROS consistently performs localized, semantically grounded modifications that preserve the original scene composition while introducing affective content. For example, EROS transforms a severed animal leg hanging in the woods into a joyful picnic scene with balloons (Row 2), or converts a vibrant flower field into a desaturated barren landscape while maintaining the underlying scene layout (Row 3).

Different baseline families exhibit distinct failure modes. Global transformation methods (AIF and NST) primarily rely on color or stylistic changes, often failing to introduce semantically meaningful affective content, as illustrated by the flower field example where only global appearance is altered (Row 3). XDream frequently generates abstract or texture-like patterns with limited semantic correspondence to the source, for example replacing a tractor scene with noise-like visual textures (Row 1). Prompt-based diffusion methods, including SDEdit, BlipDiff, and CtrlNet, produce visually realistic edits but often replace major scene elements or modify regions unrelated to the target emotion, such as transforming the hanging-leg scene into an unrelated indoor portrait (Row 2) or replacing the original environment with a completely different landscape (Row 3). LMS generally produces semantically plausible affective content but exhibits larger semantic drift, for example replacing the original stage performer with a microphone (Row 4) or reconstructing an entirely new outdoor scene for the damaged tractor example rather than preserving the original composition (Row 1).
These qualitative observations closely mirror the quantitative findings in \textbf{Fig.~\ref{fig:fig1}E and F}, demonstrating that EROS achieves a superior balance between effective emotion induction and preservation of the original scene.

\subsubsection{EROS efficiently adapts through human-in-the-loop interaction}
\label{sec:in-lab}

\begin{figure}[!ht]
\begin{center}
\includegraphics[width=\linewidth]{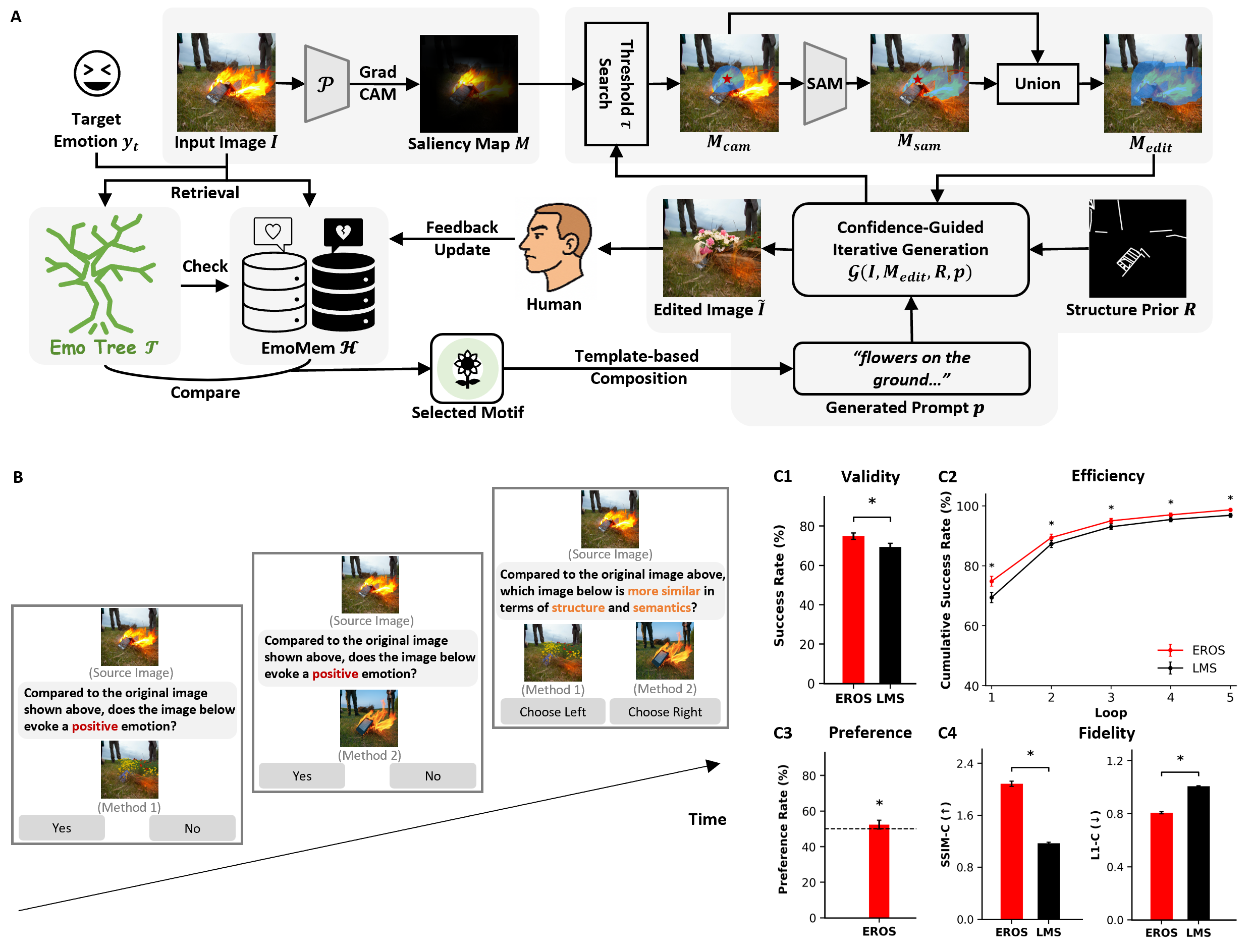}\vspace{-4mm}
\end{center}
   \caption{\footnotesize
   \textbf{EROS (Emotion-augmented geneRatiOn System), our proposed emotionally intelligent and personalized generative AI framework for emotion augmentation.
   }
   \textbf{A,} Overview of the proposed Emotion-augmented geneRatiOn System (EROS, \textbf{Sec.\ref{sec:EROS}}). Given an input image $I$ and a target valence $y_t$, EROS operationalizes emotional intelligence through five modules. \textbf{Recognition}: an emotion predictor $\mathcal{P}$ estimates the emotional valence of $I$ and generates a saliency map $M$ via Grad-CAM (\textbf{Module 1}). \textbf{Reasoning}: $M$ is thresholded to obtain $M_{\mathrm{cam}}$ and refined using SAM to produce an object-aware mask $M_{\mathrm{sam}}$, yielding the final editable region $M_{\mathrm{edit}}$ that localizes emotion-relevant image content (\textbf{Module 2}). \textbf{Control}: candidate affective rules, represented as compositional motifs consisting of concepts, attributes, actions, and scenes, are retrieved from the hierarchical EmoTree $\mathcal{T}$ and the personalized memory bank EmoMem $\mathcal{H}$. EmoTree $\mathcal{T}$ encodes symbolic affective knowledge distilled from large-scale image--emotion data (\textbf{Modules 3}), whereas EmoMem $\mathcal{H}$ stores user-specific preferences accumulated through interaction (\textbf{Modules 5}). \textbf{Generation}: the selected motif is converted into a compositional prompt $p$, and an edited image is generated as $\tilde{I}=\mathcal{G}(I,M_{\mathrm{edit}},R,p)$ using diffusion-based image editing with a structural prior $R$ that preserves scene geometry and semantics (\textbf{Module 4}). \textbf{Personalization}: human feedback updates EmoMem $\mathcal{H}$ by storing accepted and rejected motifs (\textbf{Modules 5}), enabling inference-time personalization and iterative refinement of future edits.
   \textbf{B,} Schematic illustration of the Exp-EmoInteract experiment (\textbf{Sec.\ref{sec:Exp-EmoInteract}}). In this iterative emotional-intelligence benchmark, participants evaluate affective image-editing methods across multiple interaction loops. At each loop, given a source image and a target valence, participants are presented with two candidate edits generated by EROS and a competing baseline in randomized order and independently judge whether each edited image successfully evokes the target valence relative to the source image. If both methods successfully achieve the target valence within the same loop, a pairwise comparison is conducted to determine which edited image better preserves the semantic content and structural integrity of the original scene. If only one method succeeds, the unsuccessful method proceeds to the next loop and generates a new candidate edit. If neither method succeeds, both methods continue to the next loop. This process repeats until the target valence is successfully achieved or a maximum of five loops is reached. 
    \textbf{C,} Quantitative results from Exp-EmoInteract comparing EROS with the strongest baseline, the Large Model Series (LMS, \textbf{Sec.~\ref{sec:baselines}}), which combines proprietary GPT-based large language and multimodal models for iterative affective image editing. Performance is evaluated using four complementary metrics (\textbf{Sec.\ref{sec:metric}}). \textbf{C1, Validity:} first-loop success rate in eliciting the target valence. \textbf{C2, Efficiency:} cumulative success rate across iterative editing loops. \textbf{C3, Preference:} human preference rate for EROS over LMS when both methods successfully evoke the target valence within the same loop. The dashed line indicates chance performance (50\%). \textbf{C4, Fidelity:} preservation of the semantic and structural integrity of the source image, quantified using SSIM-C (higher, better) and L1-C (lower, better).
    }
    \vspace{-5mm}
\label{fig:fig2}
\end{figure}

Practical affective image editing requires users to iteratively refine generated results through interaction rather than accepting a single edit. We therefore evaluate EROS in a personalized human-in-the-loop setting using the Exp-EmoInteract experiment (\textbf{Fig.~\ref{fig:fig2}B}; \textbf{Sec.~\ref{sec:Exp-EmoInteract}}). At each loop, participants independently evaluated edits generated by EROS and LMS (\textbf{Sec.\ref{sec:baselines}}) and indicated whether each image successfully evoked the target valence. Methods that failed to achieve the target emotion proceeded to the next editing loop until success or a maximum of five loops. We quantify interactive performance using four complementary metrics: validity, efficiency, preference, and structural fidelity (\textbf{Fig.~\ref{fig:fig2}C1--C4}; \textbf{Sec.~\ref{sec:metric}}).

EROS demonstrates superior performance throughout the interactive editing process. At the first loop, EROS achieves a significantly higher success rate than LMS (74.90\% versus 69.42\%; \textbf{Fig.~\ref{fig:fig2}C1}; Welch's two-tailed t-test, $p<10^{-3}$), indicating that EROS more reliably produces successful affective edits with minimal user interaction. More importantly, iterative feedback enables EROS to converge rapidly toward the desired emotional outcome (\textbf{Fig.~\ref{fig:fig2}C2}). The cumulative success rate increases from 74.90\% after the first loop to 89.45\% after only two loops and reaches 98.71\% by the fifth loop. Compared with LMS, EROS consistently achieves higher cumulative success rates across all interaction loops (all $p<10^{-3}$), with the largest advantage observed during the early loops. These results demonstrate that EROS not only produces better initial edits but also requires fewer rounds of interaction to satisfy users.

When both methods successfully evoked the target emotion within the same loop, participants still exhibited a significant preference for EROS over LMS (52.39\% versus the 50\% chance level; \textbf{Fig.~\ref{fig:fig2}C3}; Welch's two-tailed t-test, $p<10^{-3}$). Although the preference margin is modest, it indicates that EROS generates perceptually more favorable edits even after both methods have achieved successful emotion induction.

To evaluate whether iterative affective editing preserves the original scene while selectively modifying emotion-relevant content, we further compare structural fidelity using the proposed SSIM-C ($\uparrow$) and L1-C ($\downarrow$) metrics (\textbf{Sec.~\ref{sec:metric}}). EROS achieves substantially higher SSIM-C (2.09 versus 1.17) and lower L1-C (0.81 versus 1.01) than LMS (\textbf{Fig.~\ref{fig:fig2}C4}; both $p<10^{-3}$), indicating that its edits remain more localized to emotion-relevant regions while preserving the semantic content and structural integrity of the original image. Together, these results demonstrate that EROS not only converges more rapidly through human-in-the-loop interaction but also produces higher-quality affective edits with better structural preservation than large multimodal models.

Representative examples further illustrate the interactive editing process of EROS (\textbf{Fig.~\ref{fig:S3}}). In \textbf{Fig.~\ref{fig:S3}A1}, EROS successfully induces the target negative valence in the first loop by retrieving the motif \textit{desolate carnival} and introducing localized, semantically coherent modifications to the emotion-relevant region, whereas LMS fails to evoke the target emotion and requires an additional editing loop. In \textbf{Fig.~\ref{fig:S3}A2}, both methods successfully induce the target positive valence in the first loop. EROS is nevertheless preferred because the retrieved motif (\textit{small dog}) is integrated while better preserving the original scene structure and semantic content, whereas LMS introduces larger semantic changes. 
These examples illustrate that explicit motif retrieval enables both faster convergence during interaction and more faithful affective editing once the target emotion has been achieved.

\subsection{Dissecting the Emotional Intelligence of EROS}

Having established that EROS produces superior affective image edits, we next investigate the mechanisms underlying its performance. Specifically, we evaluate the five core capabilities of emotional intelligence implemented by EROS (\textbf{Fig.~\ref{fig:fig1}B}), namely emotion recognition, emotion understanding, affective reasoning, controllable emotion generation, and personalization. Together, these analyses provide insight into how each component contributes to the overall editing performance.

\subsubsection{EROS accurately predicts image emotion valence}
\label{sec:res_P}
We first evaluate whether the emotion predictor (\textbf{Sec.~\ref{sec:module1_recog}}) can reliably infer the overall image valence. As shown in \textbf{Fig.~\ref{fig:fig3}A}, the confusion matrix demonstrates high classification accuracy for both positive and negative images. Positive images are correctly classified at a rate of $97.0\%$, whereas negative images achieve $87.0\%$ accuracy, with low cross-class confusion (3.0\% positive-to-negative and 13.0\% negative-to-positive). These results indicate that EROS provides reliable image-level emotion recognition, forming the basis for subsequent localization of emotion-relevant regions.

\subsubsection{EROS identifies emotion-relevant regions aligned with human perception}
\label{sec:res_EmoRegion}
\begin{figure}[!ht]
\begin{center}
\includegraphics[width=\linewidth]{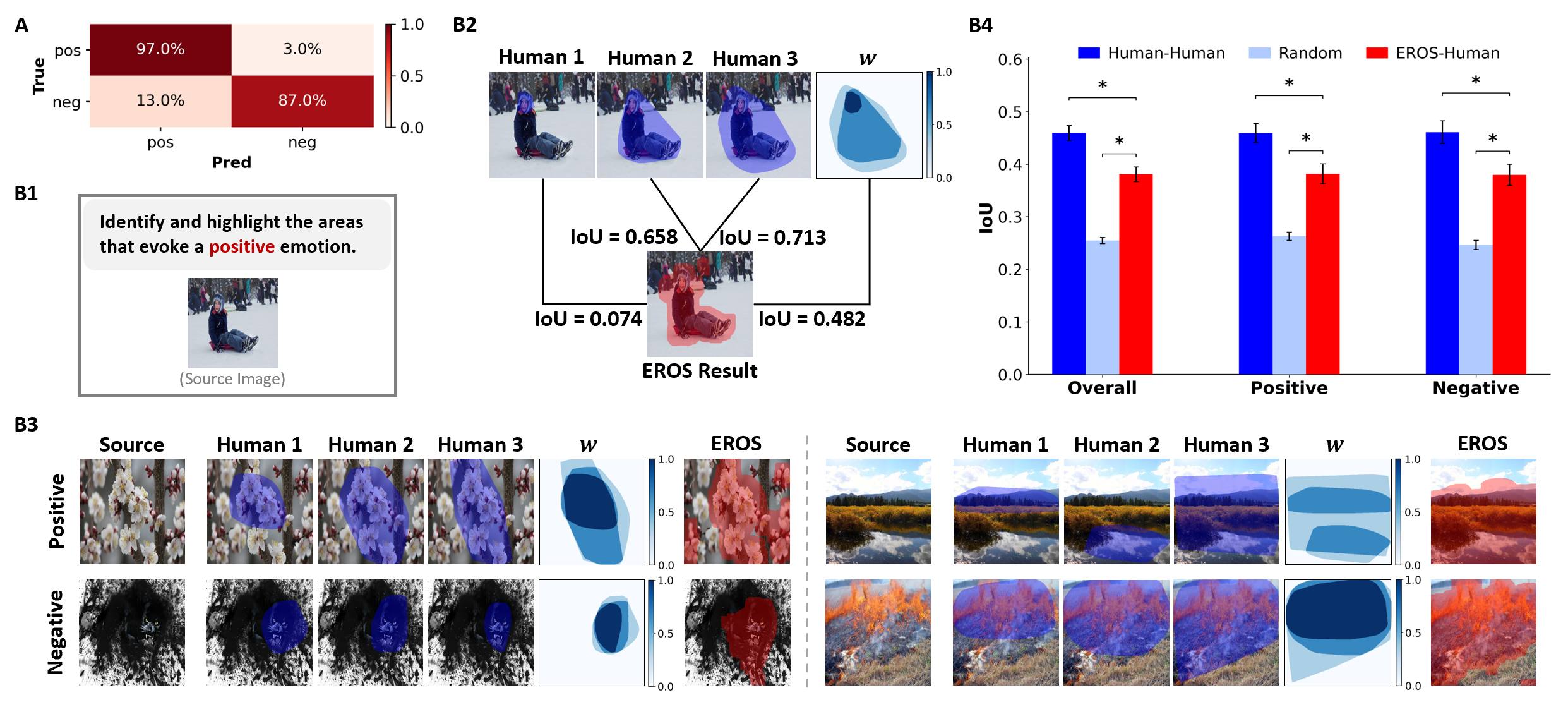}\vspace{-4mm}
\end{center}
   \caption{\footnotesize
    \textbf{Benchmarking EROS on emotion recognition, understanding, and reasoning.}
    \textbf{A,} Valence recognition performance of EROS. Confusion matrix showing the probability of predicting positive and negative valence classes. Rows correspond to ground-truth labels and columns correspond to model predictions. Color indicates prediction probability.     
    \textbf{B1,} Schematic illustration of the Exp-EmoRegion experiment (\textbf{Sec.\ref{sec:Exp-EmoRegion}}). Given an image and its emotional valence, participants identify and outline image regions that they perceive as contributing most strongly to the target emotion.    
    \textbf{B2},
    Example emotion-relevant regions identified by EROS and human observers. Human annotations from three participants are shown in blue and EROS predictions in red. The aggregated human annotation map $w$, obtained by combining annotation masks across human annotators, is shown on the right. Color intensity indicates the normalized agreement probability across human observers. Intersection-over-Union (IoU) quantifies the overlap between EROS-predicted and human-annotated regions (\textbf{Sec.\ref{sec:metric}}).    
    \textbf{B3,} Additional examples of emotion-relevant region localization. Columns 2--5 show annotations from three individual participants and the aggregated human map, respectively, overlaid on the source image (column 1). EROS predictions are shown in red and human annotations in blue, following the same convention as in B2.
    \textbf{B4,} Quantitative results from Exp-EmoRegion. IoU scores are reported for overall (left), positive-valence (middle), and negative-valence (right) conditions. Human--human agreement (blue), EROS--human agreement (red), and a random baseline (light blue) are shown for comparison. The random baseline is computed by randomly pairing human annotations from different images before calculating IoU.
    }
    \vspace{-5mm}
\label{fig:fig3}
\end{figure}

We next evaluate whether EROS correctly localizes the image regions responsible for human emotional perception (\textbf{Sec.~\ref{sec:Exp-EmoRegion}}). Human participants annotated regions that evoked the target emotion (\textbf{Fig.~\ref{fig:fig3}B1}), and the predicted masks by EROS were compared with human annotations using Intersection-over-Union (IoU, \textbf{Sec.\ref{sec:metric}}).

Representative annotations from three human participants are shown in \textbf{Fig.~\ref{fig:fig3}B2}. Although individual annotators differ in the precise spatial boundaries of their masks, they consistently identify the same semantically meaningful region (e.g., the boy in the center) responsible for the perceived emotion. We therefore aggregate the human annotations into a probabilistic weight map $w$, where higher values indicate stronger agreement across annotators. EROS closely matches these consensus regions by highlighting the entire boy, suggesting that it captures the core emotion-relevant content despite natural variability in human annotations.
Additional qualitative examples are presented in \textbf{Fig.~\ref{fig:fig3}B3}. These examples demonstrate that emotion-relevant regions vary substantially across scenes, ranging from compact foreground objects (e.g., flowers), to localized background regions (e.g., lake reflections), to large scene-level structures (e.g., wildfire). Across all examples, EROS consistently identifies regions that closely align with the aggregated human weight maps $w$, indicating robust localization across diverse image content and spatial scales.

We further quantify localization performance using three reference baselines (\textbf{Fig.~\ref{fig:fig3}B4}). The first, \emph{Human--Human}, serves as the upper bound and is computed as the average pairwise IoU between all pairs of human annotations for the same image. It represents the expected level of agreement achievable between independent human annotators, reflecting the intrinsic ambiguity of localizing emotion-relevant regions. The second, \emph{Random}, serves as the chance-level lower bound and is computed by pairing the human annotation of each image with randomly sampled human annotations from different images. Finally, \emph{EROS--Human} measures the average IoU between the region predicted by EROS and all human annotations for the corresponding image, quantifying how closely EROS aligns with human perception.
Human--Human achieves an overall IoU of $0.46 \pm 0.01$, whereas Random yields a substantially lower IoU of $0.25 \pm 0.006$, reflecting chance-level overlap. EROS--Human achieves an overall IoU of $0.38 \pm 0.01$, significantly exceeding the Random baseline across all conditions and approaching the Human--Human upper bound (all Welch's two-tailed t-tests, $p<10^{-3}$). Similar trends are observed for both positive- and negative-valence subsets. Together, these results demonstrate that EROS reliably identifies emotion-relevant regions that closely match the consensus of human annotators despite the intrinsic variability of human annotations.

\subsubsection{EROS organizes affective knowledge into an interpretable EmoTree}
\label{sec:res_tree_prompt}

EROS organizes affective knowledge into a hierarchical symbolic structure, termed the \textit{EmoTree}, according to semantic similarity and emotional valence (\textbf{Sec.~\ref{sec:module3_EmoTree}}; construction pipeline in \textbf{Fig.~\ref{fig:S4}A}). Rather than representing emotions as isolated object labels, the EmoTree links semantically related concepts to compositional affective motifs consisting of concepts, attributes, actions, and scene contexts, providing an interpretable representation for affective reasoning and image editing. Importantly, unlike traditional affective knowledge bases that rely on manually curated symbolic associations, the EmoTree is constructed automatically by mining large-scale image--emotion data, enabling scalable and data-driven acquisition of affective knowledge.

\begin{figure}[!ht]
\begin{center}
\includegraphics[width=\linewidth]{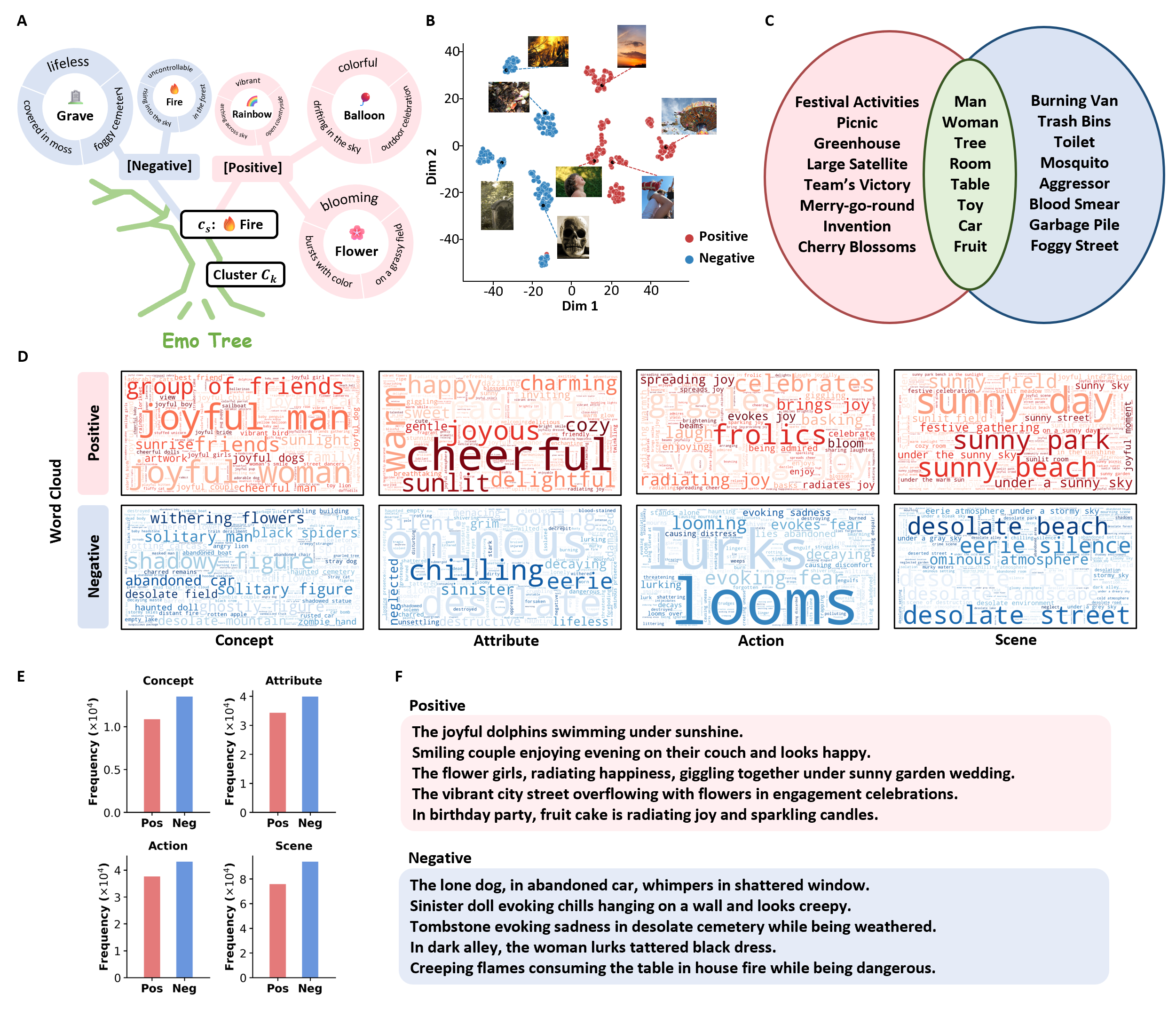}\vspace{-4mm}
\end{center}
   \caption{\footnotesize
    \textbf{Constructing the EmoTree from large-scale image--emotion dataset: a symbolic repository of affective rules for emotion-aware image editing.}
    \textbf{A},
    Schematic illustration of the EmoTree $\mathcal{T}$ (\textbf{Sec.\ref{sec:module3_EmoTree}}). The hierarchy organizes affective knowledge into semantic clusters $C_k$, source concepts $c_s$, target valences, and target motifs. Given a source concept (e.g., \textit{fire}), separate branches encode candidate transformations toward positive or negative emotional outcomes. Each target motif is represented compositionally by a target concept $c_t$ (e.g., \textit{grave}), together with its associated attribute, action, and scene context, forming an interpretable affective rule for emotion-aware image editing.   
    \textbf{B,} Visualization of semantic clusters derived from the large-scale EmoSet dataset~\cite{yang2023emoset}. Images are represented using CLIP embeddings~\cite{radford2021CLIP} and grouped according to semantic similarity through unsupervised clustering (\textbf{Sec.\ref{sec:module3_EmoTree}}). For visualization, embeddings of four selected clusters from each valence are projected into two dimensions using PCA followed by t-SNE. Each point represents an image, with colors indicating emotional valence (red: positive; blue: negative). Representative images from the discovered clusters are overlaid in the embedding space, illustrating the semantic coherence and diversity of the clustered image groups.
    \textbf{C},
    Example source concepts $c_s$ associated with positive, neutral, and negative valence. Positive concepts (red) and negative concepts (blue) are extracted from image descriptions and organized into separate concept vocabularies. Concepts appearing in both valence groups form a neutral set (green), reflecting context-dependent emotional interpretations.
    \textbf{D},
    Word clouds of target motifs stored in the EmoTree. Motifs are represented compositionally by four components: concepts, attributes, actions, and scene contexts. Red and blue word clouds correspond to motifs associated with positive and negative valence, respectively. Word size reflects occurrence frequency across the EmoTree.
    \textbf{E},
    Frequency ($\times 10^4$) of unique motif components across positive (red, Pos) and negative (blue, Neg) valence branches of the EmoTree. Statistics are shown separately for concepts, attributes, actions, and scene contexts.
    \textbf{F,} Representative affective rules extracted from the EmoTree. Example positive (red box) and negative (blue box) motif descriptions illustrate how concepts, attributes, actions, and scene contexts in motifs are combined to form interpretable affective editing rules that guide emotion-aware image editing. See \textbf{Sec.~\ref{sec:module4_gen}} for implementation details.
    }
    \vspace{-5mm}
\label{fig:fig4}
\end{figure}

As illustrated in \textbf{Fig.~\ref{fig:fig4}A}, a source concept can give rise to distinct affective interpretations depending on the target valence. For example, the source concept \textit{fire} is associated with negative concepts such as \textit{grave} and positive concepts such as \textit{rainbow}, \textit{balloon}, and \textit{flower}. Each target concept is further expanded into attributes, actions, and scene contexts, illustrating how emotional meaning is represented through structured visual motifs rather than individual objects.

To evaluate whether the EmoTree preserves semantic organization, we extract CLIP embeddings from EmoSet images \cite{yang2023emoset} (\textbf{Sec.~\ref{sec:Emoset}}) and perform hierarchical clustering based on pairwise image cosine similarity (\textbf{Sec.~\ref{sec:module3_EmoTree}}). For visualization, the embeddings are projected into two dimensions using PCA followed by t-SNE. \textbf{Fig.~\ref{fig:fig4}B} shows representative semantic clusters together with their centroid-nearest prototype images. Images naturally form compact clusters with strong internal visual consistency, spanning diverse visual themes such as natural landscapes, human portraits, amusement parks, fire, skulls, graves, and garbage. These semantically coherent neighborhoods enable EROS to retrieve target-valence motifs from visually similar source concepts while preserving semantic consistency with the input image.

Having established the semantic organization of the EmoTree, we next examine the affective concepts discovered within each valence branch. As shown in \textbf{Fig.~\ref{fig:fig4}C}, positive concepts are dominated by social activities, celebrations, and natural environments (e.g., \textit{festival activities}, \textit{picnic}, and \textit{cherry blossoms}), whereas negative concepts emphasize threats, deterioration, and hazardous environments (e.g., \textit{burning van}, \textit{trash bins}, and \textit{blood smear}). A smaller subset of concepts (e.g., \textit{man}, \textit{woman}, \textit{tree}, and \textit{fruit}) appears in both branches, indicating that object identity alone is often insufficient to determine emotional valence. Instead, emotional meaning emerges through contextual composition, with shared concepts acquiring different affective meanings through their associated attributes, actions, and scene contexts.

This compositional organization is further illustrated by the valence-specific vocabularies in \textbf{Fig.~\ref{fig:fig4}D}. After removing symbols shared between the two branches, the remaining concepts, attributes, actions, and scene contexts reveal distinct affective vocabularies, where word size reflects frequency. Positive motifs are characterized by people, friendship, and social activities, together with attributes such as \textit{cheerful} and \textit{joyous}, actions including \textit{celebrates} and \textit{radiates}, and scene contexts such as \textit{sunny beach} and \textit{sunny park}. In contrast, negative motifs emphasize threatening or deteriorating concepts, attributes such as \textit{eerie} and \textit{chilling}, actions including \textit{lurks} and \textit{looms}, and scene contexts such as \textit{desolate street} and \textit{eerie silence}.

Beyond individual examples, the global motif distribution exhibits a clear asymmetry between positive and negative affect. At the concept level, positive motifs are dominated by a relatively small set of recurring human-centered concepts, whereas negative concepts are distributed across a broader range of threatening figures, deteriorating environments, and ominous objects, as further confirmed by the top-20 concept frequencies in \textbf{Fig.~\ref{fig:S4}D}. This asymmetry is quantified in \textbf{Fig.~\ref{fig:fig4}E}, where the negative branch consistently contains more unique concepts, attributes, actions, and scene contexts than the positive branch. These findings suggest that positive affect converges onto a relatively stable semantic core, whereas negative affect can be elicited through a much wider variety of visual cues, echoing Tolstoy's observation that ``all happy families are alike; each unhappy family is unhappy in its own way''~\cite{tolstoy2016anna}.

Finally, the discovered motifs are instantiated as executable affective editing rules. EROS converts each retrieved motif into a natural-language editing instruction using predefined templates (\textbf{Sec.~\ref{sec:module3_EmoTree}}), with representative examples shown in \textbf{Fig.~\ref{fig:fig4}F}. Although these template-based prompts may exhibit minor grammatical imperfections, they convey the intended emotion unambiguously. Importantly, emotional valence emerges from the interaction among motif components rather than from individual concepts alone. For example, the concept \textit{dog} becomes strongly negative when combined with \textit{abandoned car} and \textit{shattered window}, whereas \textit{fruit cake} evokes positive affect when paired with \textit{birthday party} and \textit{sparkling candles}. Likewise, concepts with inherent emotional meaning can be further reinforced through contextual composition, such as \textit{creeping flames} embedded in \textit{consuming the table in a house fire}. These structured interactions bridge symbolic affective reasoning and controllable image generation through interpretable, executable editing rules.

\subsubsection{Generated affective rules by EROS reliably convey the target emotions}
\label{sec:res_emoprompt}
We next evaluate whether the textual prompts generated from the EmoTree faithfully encode the intended emotional valence (\textbf{Fig.~\ref{fig:fig5}A1}). Participants were asked to judge whether each generated sentence described a \textit{positive} or \textit{negative} scene (\textbf{Sec.~\ref{sec:Exp-EmoPrompt}}), thereby directly assessing whether the symbolic affective rules are interpretable without visual context. We quantify this behavior using the \emph{alignment score} (\textbf{Sec.~\ref{sec:metric}}), defined as the percentage of prompts for which the emotional valence intended by EROS agrees with human judgments.

As shown in \textbf{Fig.~\ref{fig:fig5}A2}, the generated prompts achieve an overall alignment score of $90.32\%\pm0.45\%$, substantially exceeding chance level (50\%; Welch's two-tailed t-test, $p<10^{-3}$). Similar performance is observed for both positive ($93.00\%\pm0.54\%$) and negative ($87.65\%\pm0.71\%$) prompts (both $p<10^{-3}$), demonstrating that the affective rules generated by EROS consistently convey the intended emotional valence.
These results provide direct evidence that the affective rules stored in the EmoTree are semantically meaningful rather than arbitrary symbolic combinations. Although the prompts are generated from simple templates and may not always be grammatically perfect, human observers reliably recover the intended emotional meaning. This demonstrates that emotional valence emerges from the compositional interaction among concepts, attributes, actions, and scene contexts, rather than from isolated visual concepts alone. Independent of the image generation model in \textbf{Sec.\ref{sec:module4_gen}}, the EmoTree provides an interpretable and modular bridge between structured affective reasoning and controllable image editing.


\subsubsection{Affective rules retrieved from the EmoTree produce more effective image edits}
\label{sec:res_emorule_stage1}

\begin{figure}[!ht]
\begin{center}
\includegraphics[width=0.9\linewidth]{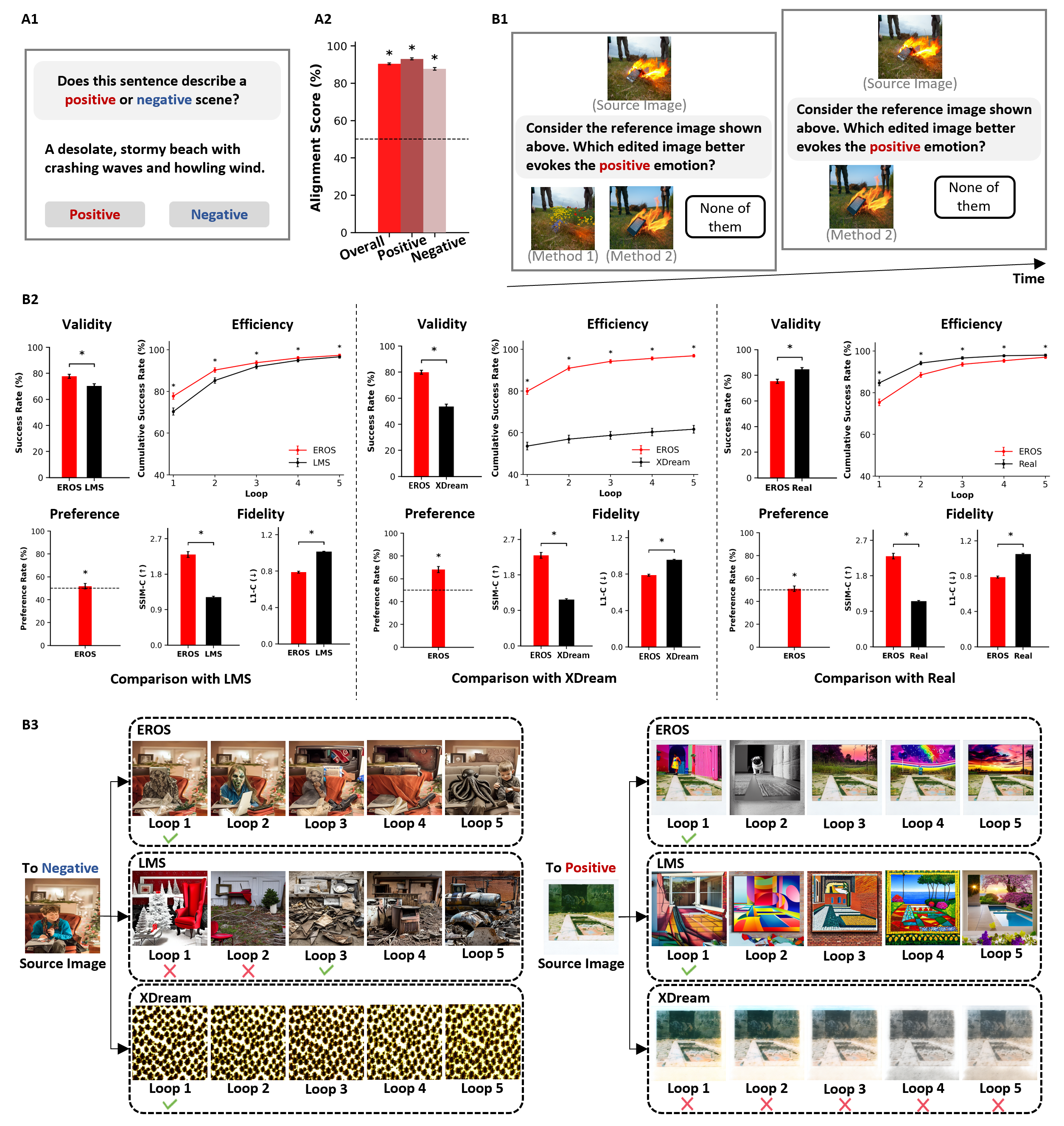}\vspace{-4mm}
\end{center}
\caption{\footnotesize
\textbf{Benchmarking EROS on iterative emotion regulation.}
\textbf{A1,} Schematic illustration of the Exp-EmoPrompt experiment (\textbf{Sec.\ref{sec:Exp-EmoPrompt}}). Participants are presented with a textual description generated from an EmoTree motif and determine whether it conveys positive or negative emotional valence.
\textbf{A2,} Results from Exp-EmoPrompt. Alignment scores quantify the agreement between the intended valence of EmoTree motifs and human judgments. Scores are reported for overall, positive-valence, and negative-valence conditions. The dashed line indicates chance performance (50\%). 
    \textbf{B1}, Schematic illustrations of the Exp-EmoRule experiment (\textbf{Sec.\ref{sec:Exp-EmoRule}, Stage 1}).
    In the first stage, participants compare edited images generated by different methods and select the image that best evokes the target valence, with the option to reject both images. Methods that fail to evoke the target valence proceed to subsequent editing loops. Unlike the Exp-EmoInteract experiment, all personalization mechanisms of all the methods are disabled in this experiment, allowing the affective rules encoded in EmoTree $\mathcal{T}$ of EROS to be evaluated independently of user-specific preferences. The same setting is applied to all baseline methods. 
    \textbf{B2,} Quantitative results from Exp-EmoRule comparing EROS against three baseline methods LMS, XDream, and Real. Performance is evaluated using four complementary metrics: \textit{Validity} (first-loop success rate), \textit{Efficiency} (cumulative success rate across iterative editing loops), \textit{Preference} (human preference rate when competing methods successfully achieve the target valence at the same loop), and \textit{Fidelity} (structural preservation measured by SSIM-C and L1-C). See \textbf{Sec.\ref{sec:metric}} for metric introductions.
    \textbf{B3,} Representative example trials of iterative affective image editing without personalization mechanisms for negative- and positive-valence targets. Given the same source image and a target valence, EROS and competing baseline methods progressively refine image content across editing loops to achieve the desired emotional outcome. Green check marks and red crosses indicate whether participants judged the edited image at the corresponding loop to have successfully evoked the target valence. For visualization purposes, image sequences are shown for all five loops. In the actual experiment, once an edited image successfully achieved the target valence (green check mark), subsequent loops were not presented to participants; the remaining images illustrate the candidate edits that would have been generated had the iterative process continued.
}    
\vspace{-5mm}
\label{fig:fig5}
\end{figure}

We verify that the affective rules learned by the EmoTree are not only interpretable at the symbolic level but also directly useful for image synthesis. To evaluate affective editing independently of real-time interaction, we conduct a large-scale online experiment using pre-generated multi-loop editing results (\textbf{Sec.~\ref{sec:Exp-EmoRule}}). As illustrated in \textbf{Fig.~\ref{fig:fig5}B1}, participants compare two candidate edits (EROS-NoMem versus a baseline) at each loop and select the image that better evokes the target emotion relative to the source. See \textbf{Sec.~\ref{sec:EROS_Variants}} for the introduction to EROS-NoMem. If neither image successfully evokes the target emotion, participants select ``None of them,'' and evaluation proceeds to the next retrieved affective rule for up to five loops. Performance is quantified using four complementary metrics—validity, efficiency, preference, and fidelity (\textbf{Sec.~\ref{sec:metric}}).

EROS consistently retrieves more effective affective rules than existing image editing methods (\textbf{Fig.~\ref{fig:fig5}B2}). At the first loop, EROS achieves significantly higher validity than LMS ($77.75\%$ vs.\ $70.34\%$, Welch's two-tailed t-test, $p<10^{-3}$) and substantially outperforms XDream ($79.87\%$ vs.\ $53.59\%$, $p<10^{-3}$), demonstrating that the highest-ranked motifs retrieved from the EmoTree are more likely to evoke the intended emotional response on the first attempt. Compared with the Real upper bound, which directly retrieves emotionally matched images from the EmoSet dataset \cite{yang2023emoset}, EROS exhibits slightly lower initial validity ($75.36\%$ vs.\ $84.69\%$, $p<10^{-3}$). Nevertheless, EROS already approaches the performance of this upper bound despite the substantially more challenging task of editing the source image while preserving its semantic content and structure.

This advantage persists throughout subsequent loops. Against LMS, the cumulative success rate of EROS increases from $77.75\%$ at Loop~1 to $90.23\%$ at Loop~2 and reaches $97.30\%$ by Loop~5, remaining significantly higher than LMS at every loop (all loops, Welch's two-tailed t-test, $p<10^{-3}$). The improvement is even more pronounced against XDream, where EROS maintains a large performance margin throughout the entire retrieval trajectory (all loops, Welch's two-tailed t-test, $p<10^{-3}$). Although EROS starts below the Real upper bound, the gap rapidly narrows as the loop number increases (Loop~5: $97.02\%$ vs.\ $98.00\%$, all $p<10^{-3}$), suggesting that searching multiple affective motifs within the EmoTree progressively approaches the emotional effectiveness of naturalistic images with the target valence while remaining constrained by the semantic content and structure of the source image.

When both methods successfully evoke the target emotion at the same loop, participants consistently prefer the images generated by EROS. The advantage is modest but significant over LMS ($51.81\%$ vs.\ chance, $p<10^{-3}$), suggesting that the interpretable, motif-guided editing strategy of EROS provides a small but consistent perceptual advantage even when both methods successfully achieve the target emotion. The margin becomes substantially larger against XDream ($68.08\%$, $p<10^{-3}$), reflecting the superior semantic quality of EROS edits. EROS is also slightly preferred over the Real upper bound ($51.05\%$, $p<10^{-3}$), indicating that emotionally aligned edits preserving the source image are perceived as more appropriate than retrieved naturalistic images that, although affectively correct, often differ substantially in semantic content and scene context.

The superior perceptual quality of edited images by EROS is accompanied by substantially higher structural fidelity. Across all comparisons, EROS consistently achieves higher SSIM-C and lower L1-C than competing approaches. Against LMS, EROS achieves an SSIM-C of $2.3$ versus $1.2$ and an L1-C of $0.8$ versus $1.0$ (both $p<10^{-3}$). Similar improvements are observed against XDream (SSIM-C: $2.3$ vs.\ $1.2$; L1-C: $0.79$ vs.\ $0.96$) and even against the Real upper bound (SSIM-C: $2.3$ vs.\ $1.2$; L1-C: $0.79$ vs.\ $1.05$; all $p<10^{-3}$). These results indicate that the affective rules retrieved from the EmoTree guide localized, semantically meaningful modifications rather than large, uncontrolled changes, enabling EROS to achieve strong affect modulation while preserving the structural integrity of the source image.

Representative iterative affective editing trajectories further illustrate the behavior of EROS compared with LMS and XDream (\textbf{Fig.~\ref{fig:fig5}B3}). Starting from a positive scene, EROS successfully achieves negative affect modulation at the first retrieval loop by introducing a semantically meaningful negative concept (\textit{a ghostly figure}) into the emotion-relevant foreground while preserving the overall scene layout. Although only the accepted first-loop result would be presented during the experiment, all five retrieved motifs are shown to illustrate that the EmoTree supports multiple semantically distinct negative interventions. In contrast, LMS succeeds only after extensive scene degradation, whereas XDream relies primarily on visually disturbing textures with limited semantic correspondence to the source image.
Similarly, starting from a negative scene, EROS achieves positive affect modulation at the first loop by introducing semantically meaningful positive cues (\textit{a woman standing in front of a colorful house}) while preserving the original scene structure. Here, the primary modification occurs in the background rather than the foreground, demonstrating that EROS adapts the editing region according to the localized source of emotional content. Although LMS also succeeds at the first loop, it relies on highly stylized transformations that substantially alter the appearance of the original image, whereas XDream fails to generate convincing positive edits across all loops.

\textbf{Fig.~\ref{fig:S2}} further demonstrates the flexibility of EROS in generating diverse affective edits from a single source image. By retrieving different motifs from the EmoTree, EROS produces multiple semantically coherent outputs under the same target valence. For example, positive edits may emphasize natural scenery, social activities, or floral elements, whereas negative edits introduce deterioration, isolation, or hazardous content, illustrating that affective image editing is inherently a one-to-many problem rather than a deterministic mapping.
The examples also highlight the versatility of EROS across cross-valence and same-valence editing (\textbf{Sec.~\ref{sec:module4_gen}}). Cross-valence editing replaces emotion-relevant cues to reverse the perceived emotion while preserving the overall scene semantics. In contrast, same-valence editing strengthens the existing emotion by enriching its surrounding context while leaving the primary emotion-evoking regions unchanged; for example, adding a joyful crowd to the forest scene further enhances its positive affect (Row 1). 
Likewise, for everyday scenes with subtle affective cues, EROS strengthens the perceived emotion by introducing additional positive elements, such as adding potted plants to a study table to create a warmer and more inviting atmosphere (Row 3). Together, these examples demonstrate that EROS supports both emotion reversal and emotion enhancement within a unified affective image editing framework.

\subsubsection{EROS acquires interpretable personalized affective memories}
\label{sec:res_emorule_stage23}
\begin{figure}[!ht]
\begin{center}
\includegraphics[width=\linewidth]{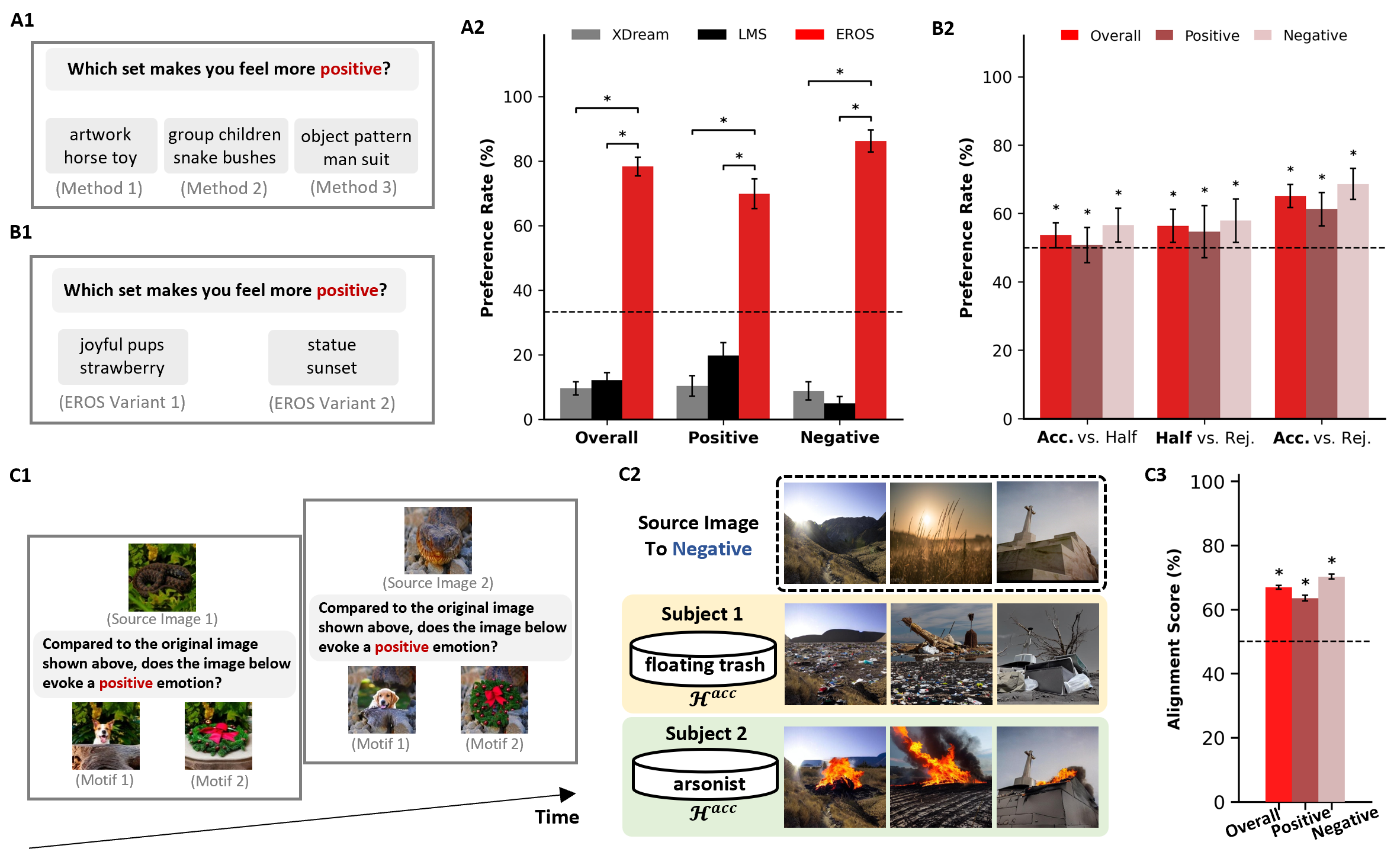}\vspace{-4mm}
\end{center}
   \caption{\footnotesize
    \textbf{
    Benchmarking personalized affective profiles in EROS.}
    \textbf{A1}, Schematic illustration of Stage 2 in the Exp-EmoRule experiment (\textbf{Sec.\ref{sec:Exp-EmoRule}}, \textbf{Stage 2}). Participants are presented with three sets of concepts retrieved from the personalized memories of different methods and select the concept set that best matches their personal preference for the target valence.
    \textbf{A2},
    Results from Stage 2 of Exp-EmoRule. Preference rates (\textbf{Sec.\ref{sec:metric}}) compare the quality of personalized affective profiles learned by EROS (red), LMS (black), and XDream (grey) across overall, positive-valence, and negative-valence conditions. The dashed line indicates chance performance (33.3\%).
    \textbf{B1,} Schematic illustration of Stage 3 in Exp-EmoRule (\textbf{Sec.\ref{sec:Exp-EmoRule}}, \textbf{Stage 3}). For a given target valence, participants compare concept sets containing different proportions of accepted and rejected concepts retrieved from the personalized memory bank EmoMem $\mathcal{H}$. Importantly, both accepted and rejected concepts correspond to the same target valence; they differ only in user preference, with accepted concepts representing motifs that participants consistently preferred over alternative affective interventions.
    Three concept sets are constructed: an all-accepted set (100\% concepts sampled from $\mathcal{H}^{acc}_{con}$), a mixed set (50\% concepts from $\mathcal{H}^{acc}_{con}$ and 50\% from $\mathcal{H}^{rej}_{con}$), and an all-rejected set (100\% concepts sampled from $\mathcal{H}^{rej}_{con}$). Participants perform three pairwise comparisons—All Accepted versus Mixed (Acc. vs. Half), Mixed versus Rejected (Half vs. Rej.), and All Accepted versus Rejected (Acc. vs. Rej.)—and, for each comparison, select the concept set they would prefer for evoking the target valence.
    \textbf{B2},
    Preference rates (\textbf{Sec.\ref{sec:metric}}) for accepted concept sets across the three pairwise comparisons described in \textbf{B1}. Each bar represents the probability that participants selected the bolded concept set in each comparison (Acc.\ vs.\ Half, Half vs.\ Rej., and Acc.\ vs.\ Rej.). The dashed line indicates chance level (50\%). 
    \textbf{C1,} Schematic illustration of the Exp-EmoMotif experiment (\textbf{Sec.\ref{sec:Exp-EmoMotif}}). Participants first select the preferred edit generated using one of two motifs from EmoTree $\mathcal{T}$ of EROS for a source image. The same pair of motifs is subsequently applied to a semantically similar image, allowing cross-scene consistency of motif preferences to be evaluated.
    \textbf{C2},
    Representative examples of cross-scene motif consistency. 
    For the same target negative valence, two participants were each presented with two candidate edits for three semantically related source images (Row 1). Subject 1 consistently preferred motifs involving \textit{floating trash} (Row 2), whereas Subject 2 consistently preferred motifs involving an \textit{arsonist} (Row 3). These personalized motif preferences were subsequently stored in their respective accepted memory banks, $\mathcal{H}^{acc}$.
    \textbf{C3},
    Quantitative results from Exp-EmoMotif. Alignment scores measure the consistency of motif preferences across semantically related image pairs for overall, positive-valence, and negative-valence conditions. The dashed line indicates chance performance (50\%).
    }
    \vspace{-5mm}
\label{fig:fig6}
\end{figure}

EROS personalizes affective image editing through an explicit symbolic memory, termed EmoMem (\textbf{Sec.~\ref{sec:module5_Personalization}}).
Rather than encoding user preferences implicitly in model parameters, EmoMem stores structured affective motifs that users have either accepted or rejected during interaction. Specifically, the accepted memory $\mathcal{H}^{acc}$ contains motifs preferred by the user, whereas the rejected memory $\mathcal{H}^{rej}$ stores motifs that failed to evoke the intended emotional response. These symbolic memories are subsequently retrieved to guide future affective reasoning and image editing. We evaluate EmoMem from three complementary perspectives: whether the stored concepts align with human preferences (\textbf{Fig.~\ref{fig:fig6}A}), whether accepted and rejected memories form discriminative affective representations (\textbf{Fig.~\ref{fig:fig6}B}), and whether the learned preferences generalize across semantically related scenes (\textbf{Fig.~\ref{fig:fig6}C}).

We first evaluate whether the concepts stored in EmoMem capture participant-specific affective preferences (\textbf{Sec.~\ref{sec:Exp-EmoRule}}, Stage~2). In each trial, participants select the concept set that best matches the target emotional valence (\textbf{Fig.~\ref{fig:fig6}A1}). Preference rates are computed as the proportion of selections for each method, with chance corresponding to random selection among the three candidate concept sets (33.3\%).
As shown in \textbf{Fig.~\ref{fig:fig6}A2}, EROS is overwhelmingly preferred over both LMS and XDream, achieving an overall preference rate of $78.31\%$, compared with $12.07\%$ for LMS and $9.58\%$ for XDream (both Welch's two-tailed t-tests, $p<10^{-3}$). The advantage is consistent for both positive ($69.86\%$) and negative ($86.23\%$) valence conditions. These results demonstrate that the symbolic concepts retrieved by EROS align substantially better with individual affective judgments than concepts extracted from existing editing methods. Unlike EROS, whose concepts are retrieved directly from the structured motifs stored in the EmoTree, baseline methods derive concepts post hoc through image captioning and linguistic parsing, which often capture incidental visual content rather than the true drivers of emotional perception.

Having established that the learned concepts align with human preferences, we next examine whether accepted and rejected memories form coherent personalized representations (\textbf{Sec.~\ref{sec:Exp-EmoRule}}, Stage~3). Participants compare concept sets constructed from different proportions of accepted ($\mathcal{H}^{acc}$) and rejected ($\mathcal{H}^{rej}$) memories (\textbf{Fig.~\ref{fig:fig6}B1}).
As shown in \textbf{Fig.~\ref{fig:fig6}B2}, human preference increases monotonically with the proportion of accepted concepts. The strongest separation occurs between fully accepted and fully rejected concept sets ($65.09\%\pm3.34$, Welch's two-tailed t-test, $p<10^{-3}$), whereas intermediate comparisons (Accepted vs.\ Half and Half; Half and Half vs.\ Rejected) produce progressively smaller but still significant differences (all $p<10^{-3}$). This monotonic ordering (Accepted $>$ Half $>$ Rejected) indicates that EmoMem successfully acquires personalized affective concepts that resonate with individual users while clearly separating preferred from rejected affective motifs. The same trend is observed for both positive and negative valence, with stronger separation for negative emotion, consistent with the greater semantic diversity of negative motifs in the EmoTree (\textbf{Sec.~\ref{sec:res_emorule_stage1}}).

\textbf{Fig.~\ref{fig:S6}A} and \textbf{Fig.~\ref{fig:S6}B} further visualize representative accepted ($\mathcal{H}^{acc}$) and rejected ($\mathcal{H}^{rej}$) memories collected in EmoMem during Exp-EmoInteract (\textbf{Sec.~\ref{sec:Exp-EmoInteract}}).
Analysis of $\mathcal{H}^{acc}$ (\textbf{Fig.~\ref{fig:S6}A}) reveals several consistent patterns in the affective motifs preferred by users. First, preferred edits often preserve the core concept while modifying its emotional interpretation through attributes, actions, and scene context. For example, the concept \textit{woman} is retained, whereas motifs such as \textit{joyful}, \textit{sits chillingly}, \textit{in a sunny field}, and \textit{in a wilted field} determine the perceived emotional valence. Second, users frequently favor transformations toward semantically related concepts with stronger affective associations, such as \textit{dog} $\rightarrow$ \textit{group of dogs} for positive editing, \textit{spider} $\rightarrow$ \textit{butterfly}, and \textit{bird} $\rightarrow$ \textit{frail bird} for negative editing. Similarly, \textit{carousel} is transformed into a \textit{desolate carnival}, illustrating that preferred edits typically remain within the same semantic neighborhood while introducing emotionally meaningful motifs.

In contrast, $\mathcal{H}^{rej}$ (\textbf{Fig.~\ref{fig:S6}B}) captures affective motifs that users consistently disfavor relative to alternative valid edits. Rejected motifs span all symbolic levels, including concepts (e.g., \textit{lion}, \textit{field}, \textit{colorful doors}, and \textit{shadows}), attributes (e.g., \textit{colorful}, \textit{cold}, \textit{happy}, and \textit{helpless}), actions (e.g., \textit{rest}, \textit{enveloping}, \textit{brightens}, and \textit{crash}), and scene contexts (e.g., \textit{on the sand}, \textit{abandoned field}, \textit{live music event}, and \textit{darkening woods}). These motifs are not necessarily ineffective at conveying the target emotion; rather, they are less preferred than competing alternatives, suggesting that users exhibit consistent preferences for certain semantic realizations of the same emotional valence.

\subsubsection{EROS captures stable personalized affective preferences across scenes}
\label{sec:res_emomotif}
Finally, we investigate whether the personalized memories stored in EmoMem generalize beyond the images from which they were acquired (\textbf{Sec.~\ref{sec:Exp-EmoMotif}}). Participants independently evaluate edits generated using the same motifs across semantically related image pairs (\textbf{Fig.~\ref{fig:fig6}C1}). Representative examples (\textbf{Fig.~\ref{fig:fig6}C2}) show that different participants consistently favor different affective motifs. For example, one participant repeatedly selects pollution-related motifs (\textit{floating trash}), whereas another consistently prefers fire-related motifs (\textit{arsonist}). Despite these differences across individuals, each participant maintains remarkably stable motif preferences across semantically related scenes.

This consistency is confirmed quantitatively in \textbf{Fig.~\ref{fig:fig6}C3}. Alignment scores are significantly above chance (50\%), reaching $66.85\%\pm0.60\%$ overall, with $63.50\%\pm0.86\%$ for positive and $70.24\%\pm0.81\%$ for negative valence (all Welch's two-tailed t-tests, $p<10^{-3}$). These results demonstrate that personalized affective preferences remain stable across semantically related scenes, enabling EmoMem to encode reusable symbolic memories that support future affective editing.

\section{Discussion}

Emotional intelligence has long been recognized as a defining component of human cognition, yet remains poorly operationalized in artificial intelligence. Existing affective image editing methods (\textbf{Sec.\ref{sec:baselines}}) are primarily designed as generative models that optimize perceptual realism or text-image alignment, rather than emotionally intelligent systems capable of recognizing emotions, reasoning about their underlying causes, adapting to individual users, and regulating emotions through interaction. In this work, we formulate personalized affective image editing as a computational framework for emotional intelligence. Specifically, we decompose emotional intelligence into five complementary capabilities: emotion recognition, emotion localization, affective reasoning, controllable emotion regulation, and personalization through memory (\textbf{Fig.~\ref{fig:fig1}B,C}). To systematically study these capabilities, we introduce standardized human psychophysics protocols (\textbf{Sec.\ref{sec:exp_all}}), quantitative evaluation metrics (\textbf{Sec.\ref{sec:metric}}), and a comprehensive benchmark comprising 11 baselines, and 33,380 human judgments collected from 296 participants across six experiments. Together, these benchmarks provide a reproducible framework for evaluating emotionally intelligent visual systems beyond conventional image generation metrics.

Our experiments further reveal several fundamental limitations of existing affective image editing approaches (\textbf{Fig.\ref{fig:fig1}G, Fig.\ref{fig:S1}, Fig.\ref{fig:fig5}B3, Fig.\ref{fig:S3}}). Global appearance transformation methods \cite{pitie2007automated,gatys2015neural,kwon2022clipstyler,weng2023affective} primarily manipulate low-level visual statistics without identifying the semantic causes of emotion. Diffusion-based editing models \cite{meng2021sdedit,zhang2023controlnet,brooks2023instructpix2pix,li2023blip} often generate emotionally plausible content but frequently hallucinate objects, replace scene layouts, or sacrifice the semantic identity of the source image. Large multimodal models \cite{hurst2024gpt,openai2025o4} demonstrate impressive semantic reasoning, particularly for positive affect, but remain limited by weak interpretability, hallucination, poor control over localized editing, and safety guardrails that constrain the generation of certain forms of negative affect. These findings suggest that emotional intelligence requires not only powerful generative models but also explicit reasoning about which visual elements should be modified, why they should be modified, and how those modifications should preserve the semantic integrity of the original scene.

To address these challenges, we propose EROS, a hybrid framework that integrates symbolic affective reasoning with deep generative models (\textbf{Sec.\ref{sec:EROS}}, \textbf{Fig.\ref{fig:fig2}A}). Rather than relying solely on latent representations, EROS explicitly identifies emotion-relevant regions, retrieves structured affective motifs from the hierarchical EmoTree, incrementally refines edits through human interaction, and stores personalized symbolic preferences in EmoMem. Across comprehensive human evaluations, compared to existing affective editing methods, EROS consistently generates edits that more effectively evoke the target emotional response while better preserving the semantic content and structural integrity of the source image than existing approaches (\textbf{Sec.~\ref{sec:pairwise}}, \textbf{Fig.~\ref{fig:fig1}D--F}). Interactive experiments further demonstrate that EROS efficiently adapts to human feedback, achieving higher validity, faster convergence, stronger user preference, and improved structural fidelity throughout the editing process (\textbf{Sec.~\ref{sec:in-lab}}, \textbf{Fig.~\ref{fig:fig2}B, C1--4}). Together, these findings suggest that affective image editing benefits from explicitly reasoning about emotion-relevant visual content rather than relying solely on end-to-end generative optimization.

More importantly, our analyses show that each component of EROS contributes a distinct aspect of emotional intelligence. The emotion recognition module reliably predicts image valence and identifies emotion-relevant regions that closely align with human annotations (\textbf{Sec.~\ref{sec:res_P}--\ref{sec:res_EmoRegion}}, \textbf{Fig.~\ref{fig:fig3}}). Building upon these perceptual representations, the automatically constructed EmoTree distills large-scale image--emotion data into interpretable symbolic affective rules whose emotional meanings are consistently understood by humans and generalize across edited scenes (\textbf{Sec.~\ref{sec:res_tree_prompt}--\ref{sec:res_emorule_stage1}}, \textbf{Fig.~\ref{fig:fig4}--\ref{fig:fig5}}). Interestingly, the learned symbolic organization also reveals an asymmetry between positive and negative emotion. Positive emotions are organized around a relatively concentrated semantic core, whereas negative emotions are supported by a broader and more diverse set of visual motifs, echoing Tolstoy's observation~\cite{tolstoy2016anna} that "all happy families are alike; each unhappy family is unhappy in its own way." Finally, EmoMem acquires accurate and explicit user-specific affective preferences that remain stable across semantically related scenes, demonstrating that personalization can be achieved through transparent symbolic memories rather than implicit parameter adaptation (\textbf{Sec.~\ref{sec:res_emorule_stage23}--\ref{sec:res_emomotif}}, \textbf{Fig.~\ref{fig:fig6}}). Collectively, these findings suggest that emotional intelligence can emerge from the integration of interpretable reasoning, structured knowledge, and human feedback.

Several important directions remain for future research. 
First, although EROS can support affective editing across a wide range of everyday scenes (\textbf{Fig.~\ref{fig:S2}}), the present work primarily focuses on cross-valence transfer. A more systematic investigation of same-valence affect enhancement and affect induction from emotionally neutral scenes toward positive or negative targets would provide a more complete understanding of emotion regulation.
Second, EROS models emotion using binary valence, whereas human affect spans a much richer taxonomy including amusement, awe, fear, anger, sadness, disgust, and surprise. Extending the EmoTree and EmoMem toward fine-grained emotional categories therefore represents an important direction for future work. More broadly, although this work focuses on affective image editing, the proposed framework is modality-independent. The combination of interpretable symbolic reasoning, explicit personalized memory, and standardized human-centered evaluation provides a foundation for developing emotionally intelligent systems across images, videos, language, and embodied agents.

Finally, while EROS enables beneficial applications such as personalized communication, assistive content generation, mental health interventions, and emotionally adaptive human--AI interaction, it also raises important ethical considerations. These capabilities could be misused for emotional manipulation, persuasion, or deceptive media, while personalization may inadvertently reinforce individual biases. As emotionally intelligent generative systems become increasingly capable of shaping human affect, ensuring transparency, user consent, user control over personalization, and responsible deployment will be essential to maximize societal benefit while mitigating potential misuse.

\section{Method}

\subsection{EmoSet Dataset}
\label{sec:Emoset}

Our framework builds upon EmoSet~\cite{yang2023emoset}, a large-scale affective image dataset containing approximately 120,000 images annotated by human participants with one of eight discrete emotion categories. To focus on emotional valence, we group these categories into four positive emotions (amusement, awe, contentment, and excitement) and four negative emotions (anger, disgust, fear, and sadness).

Importantly, EmoSet provides only image-level emotion annotations and does not contain paired image-editing examples, affective interventions, or textual instructions describing how an image should be modified to alter its emotional impact. Consequently, the dataset does not directly supervise emotional reasoning, affective image manipulation, or personalization. This setting presents a challenging testbed for emotionally intelligent AI, requiring models to infer how visual elements contribute to human emotions, reason about potential interventions, and generate emotionally targeted image modifications without access to explicit editing demonstrations. The absence of step-by-step supervision further enables us to evaluate whether such capabilities can emerge from structured affective knowledge and human feedback rather than large-scale task-specific annotations.

\subsection{EROS: Emotion-augmented geneRatiOn System}
\label{sec:EROS}

Inspired by the functional components of human emotional intelligence, we propose the \textit{Emotion-augmented geneRatiOn System} (EROS), a unified framework comprising five modules: \textit{recognition}, \textit{reasoning}, \textit{control}, \textit{generation}, and \textit{personalization}. Together, these modules enable EROS to estimate emotional responses to visual stimuli, identify their causes, reason about affective interventions, generate emotionally targeted image modifications, and adapt to individual users.

\subsubsection{Module 1 -- Recognition: Estimating Emotional Valence}
\label{sec:module1_recog}
The recognition module estimates the emotional valence of an input image $I$ using an emotion predictor $\mathcal{P}$. The predictor is implemented as a ResNet-18-based convolutional neural network~\cite{he2016resnet} with a binary classification head and is trained on EmoSet~\cite{yang2023emoset} using supervised learning.

Given an input image $I$, the predictor outputs
$
p(y \mid I)=\mathcal{P}(I),
$
where $y\in\{0,1\}$ denotes positive ($0$) or negative ($1$) emotional valence. The network is optimized using the cross-entropy loss and trained with the Adam optimizer (learning rate $1\times10^{-4}$, batch size 32, 100 epochs). All images are resized to $512\times512$ pixels during training.

The resulting predictor serves as the affective perception component of EROS, providing an estimate of the emotional response associated with an image and supplying supervisory signals for downstream reasoning and editing modules.

\subsubsection{Module 2 -- Reasoning: Identifying the Causes of Emotion}
\label{sec:module2_region}
Human emotional intelligence extends beyond recognizing emotions to understanding \emph{why} they arise. Analogously, the reasoning module seeks to identify the visual evidence responsible for an image's emotional valence. Given an image $I$ and the emotion predictor $\mathcal{P}$ from Module~1, the module localizes emotion-relevant regions and extracts the visual elements most strongly associated with the predicted affective response.

To identify these regions, we compute a valence-specific saliency map $M$ using Grad-CAM~\cite{selvaraju2017gradcam}. After retaining non-negative activations and normalizing the map, we obtain a continuous relevance map $M\in[0,1]^{H\times W}$ that highlights image regions contributing to the predicted valence.

To convert this representation into editable regions, we threshold the saliency map:
\[
M_{\text{cam}} = \mathbf{1}(M > \tau),
\]
where $\tau$ controls the spatial extent of the selected region. Higher thresholds produce more localized edits, whereas lower thresholds allow broader modifications. We employ a progressive threshold schedule $\tau\in\{0.5, 0.3, 0.1, 0\}$, gradually expanding the editable region during image editing until the desired emotional outcome is achieved.

The role of these regions depends on the editing objective. When strengthening an existing emotional valence, the localized regions identify affective cues that should be preserved or enhanced. Conversely, when transferring the image toward the opposite valence, these regions indicate visual evidence that should be altered or replaced. For example, flowers associated with positive affect may be preserved during positive-valence enhancement but modified when inducing negative affect.

Although Grad-CAM localizes emotionally relevant evidence, its boundaries do not necessarily align with coherent objects. To obtain semantically meaningful editing regions, we refine the localization using the Segment Anything Model (SAM)~\cite{kirillov2023SAM}.
Specifically, the peak activation location $(i^*, j^*) = \arg\max_{i,j} M_{\text{cam}}(i,j)$ is used as a point prompt to generate an object-level segmentation mask $M_{\mathrm{sam}}$. The final editable region is defined as:
\[
M_{\text{final}} = \max(M_{\text{cam}}, M_{\text{sam}}),
\]
where the max operation is applied element-wise. This union combines the complementary strengths of both representations: Grad-CAM identifies emotion-relevant evidence, whereas SAM provides object-level coherence. Restricting edits to $M_{\mathrm{final}}$ enables EROS to modify affective cues while preserving the broader semantic and structural content of the scene.

\subsubsection{Module 3 -- Control: Proposing Targeted Modifications}
\label{sec:module3_EmoTree}
Given a source image, its localized affective cues, and a desired target valence, the control module determines how the image should be modified to achieve the target emotional outcome. Rather than directly generating edits, the module performs affective intervention planning by identifying which visual concepts should be preserved, removed, enhanced, or replaced.

\noindent \textbf{Constructing the EmoTree $\mathcal{T}$}. 
To support this process, we introduce the \textit{EmoTree} $\mathcal{T}$, a hierarchical affective knowledge structure that encodes interpretable relationships between visual content and emotional responses. The EmoTree is automatically derived from large-scale image--emotion data by identifying recurring compositional patterns linking scene elements to emotional outcomes.

A key representation within the EmoTree is the \textit{visual motif}, defined as a compositional configuration of objects, attributes, interactions, and scene context that collectively contributes to an emotional response. Each affective rule specifies how a source concept may be transformed into a target motif under a desired emotional valence. These rules are organized as root-to-leaf paths within the EmoTree.

The EmoTree consists of four hierarchical levels:

\begin{enumerate}
\item \textbf{Semantic Cluster}: a high-level semantic category grouping visually related images;
\item \textbf{Source Concept}: the object or concept identified in the original image;
\item \textbf{Target Valence}: the desired emotional direction (positive or negative);
\item \textbf{Target Motif}: a compositional visual motif predicted to induce the target emotional response.
\end{enumerate}

By representing affective knowledge as explicit symbolic structures rather than latent parameters alone, the EmoTree provides an interpretable mechanism for emotional reasoning and intervention planning, allowing EROS to explain \emph{why} a particular image modification is expected to alter emotional perception.
Next, we describe the construction of the EmoTree at each hierarchical level. See \textbf{Fig.~\ref{fig:S4}A} for an overview of the construction pipeline.

\noindent \textbf{Level 1: Semantic Cluster.} 
Images are first organized into visually coherent semantic clusters. We begin by grouping EmoSet~\cite{yang2023emoset} images according to their human-annotated emotion categories. Within each category, images are further clustered based on visual similarity. For each image $I$, we extract a CLIP embedding 
$\mathbf{v}_I = f_{\mathrm{CLIP}}^{\mathrm{img}}(I) \in \mathbb{R}^d$, 
and perform hierarchical agglomerative clustering using pairwise cosine similarity. Let $\mathcal{C}={C_1,\dots,C_K}$ denote the resulting clusters. The number of clusters is controlled by a cosine similarity threshold $\delta$. Lower thresholds produce broader semantic groupings, whereas higher thresholds lead to over-fragmentation (\textbf{Fig.~\ref{fig:S4}B}). We adopt $\delta=0.7$, which balances intra-cluster coherence and inter-cluster diversity.

Each cluster $C_k$ stores a prototype embedding 
$\bar{\mathbf{v}}_k = \frac{1}{|C_k|} \sum_{I \in C_k} \mathbf{v}_I$, 
which is used during inference for semantic retrieval. Applying this procedure to EmoSet yields 5,712 semantic clusters spanning related objects, scenes, and contextual configurations.

\noindent \textbf{Level 2: Source Concept.}
For each image, we extract a source concept $c_s$ that serves as a semantic anchor for affective reasoning. We focus on noun phrases because they correspond to concrete entities that provide stable concept-level representations, whereas verbs and adjectives are often more context dependent.
For example, in an image depicting ``a crying child in a hospital bed,'' the noun phrase ``child'' or ``hospital bed'' provides a more stable and generalizable emotional anchor than the verb ``crying,'' which may appear in both positive (e.g., tears of joy) and negative contexts.

Given an image $I$, we first generate a caption $t$ using BLIP~\cite{li2022blip}. We then use GPT-4o~\cite{hurst2024gpt} to identify the entity most strongly associated with the image's annotated emotional valence $y_s$. Specifically, we query GPT-4o with the prompt:
\textit{``Analyze $t$. Identify the entity most responsible for $y_s$ emotion.''} The resulting noun phrase is taken as the source concept $c_s$. GPT-4o is used solely as a structured language parser to extract concepts from captions; the affective associations stored in the EmoTree are derived from image--emotion statistics in EmoSet rather than GPT-4o outputs alone.

Aggregating all source concepts across EmoSet yields two concept vocabularies based on ground-truth valence. Let $\mathcal{S}_{0}$ denote the set of positive concepts and $\mathcal{S}_{1}$ denote the set of negative concepts. These vocabularies provide candidate concepts for affective intervention during image editing.

Some concepts appear in both sets, indicating context-dependent emotional interpretations. We define the overlap as $\mathcal{S}_{\mathrm{neu}} = \mathcal{S}_{0} \cap \mathcal{S}_{1}$. To remove duplicates and consolidate semantically equivalent entries, we refine this set using GPT-4o by retaining only clearly neutral concepts, converting each concept to singular noun form, and enforcing single-word representations. The exact prompt to GPT-4o is: \textit{``Given a list of candidate concepts, retain only those that are clearly neutral, convert each concept to its singular noun form, and ensure that each entry is a single-word concept.''} The resulting vocabulary contains 3,640 positive concepts, 2,625 negative concepts, and 318 neutral concepts. Examples are shown in \textbf{Fig.~\ref{fig:fig4}C}.

\noindent \textbf{Level 3: Target Valence.}
For each source concept, EROS supports both same-valence enhancement and cross-valence transformation. Accordingly, each source concept branches into two directional subtrees corresponding to the desired emotional outcome: $y_t \in \{\text{positive}, \text{negative}\}$. This level explicitly represents the target emotional objective and guides subsequent affective interventions.

\noindent \textbf{Level 4: Target Motif.}
Given a source concept $c_s$ and a target valence $y_t$, EROS constructs a compositional visual motif that specifies how the concept should be transformed to achieve the desired emotional outcome. See the pipeline of an example motif reconstruction in \textbf{Fig.~\ref{fig:S4}C}.
Because multiple candidate target concepts may exist, exhaustive evaluation is computationally infeasible. Instead, for each image $I_i\in C_k$ with caption $t_i$, we randomly sample 50 concepts from the target-valence vocabulary $\mathcal{S}{y_t}$ to form a candidate pool $\mathcal{A}{y_t}$. GPT-4o is then used to select the concept $c_{cand}$ that is both semantically compatible with $t_i$ and likely to induce the target valence.
The text prompt to GPT-4o is:
\textit{``Given the source image description $t_i$, the target valence $y_t$, and a set of candidate concepts $\mathcal{A}_{y_t}$, select one concept that achieves the target valence while maintaining semantic coherence with $t_i$."}

For each selected candidate concept $c_{cand}$, we use GPT-4 to generate a visual motif $m=(c_t,a_t,r_t,s_t)$, where $c_t$ denotes the target concept, and $a_t$, $r_t$, and $s_t$ specify its attributes, related actions, and scene context, respectively. Conditioned on the source image description $t_i$, the target valence $y_t$, and the candidate concept $c_{cand}$, GPT-4 performs context-aware affective reasoning to construct a semantically coherent visual motif. Importantly, $c_{cand}$ serves as semantic guidance rather than a mandatory replacement. Depending on the source scene, GPT-4 may refine $c_{cand}$ into a more specific concept, replace the original emotion-related concept, or integrate $c_{cand}$ with the existing scene while preserving semantic coherence. For example, when the source image contains a \textit{skull} and the candidate concept $c_{cand}$ is \textit{flower} for positive editing, GPT-4 may generate the motif \textit{a skull wearing flowers} instead of replacing the skull entirely with flowers, thereby preserving the original scene semantics while introducing positive affective cues.
Specifically, the following query is provided to GPT-4, where the selected candidate concept $c_{cand}$ serves as semantic guidance for affective reasoning: 
\textit{``Given the source image description $t_i$, the target valence $y_t$, and the selected candidate concept $c_{cand}$, revise $t_i$ by replacing or modifying its emotion-related concept with a more specific target concept. Produce a new prompt that moves the image toward the target valence while maintaining semantic coherence. Generate a realistic visual motif by specifying the target concept, its attributes, related actions, and scene context."}
Each generated motif defines a leaf node in the EmoTree. Multiple motifs may be associated with the same source concept and target valence, reflecting the existence of multiple valid affective interventions that preserve the semantic identity of the original scene while inducing the desired emotional response.

A complete path in the EmoTree, $C_k \rightarrow c_s \rightarrow y_t \rightarrow m$, defines an affective rule that specifies how a source concept may be transformed toward a target emotional outcome through structured semantic modification.
To balance emotional effectiveness and semantic preservation, motifs sharing the same source concept and target valence are organized according to source--target concept similarity. For each motif with target concept $c_t$, we compute
\[
s(c_s, c_t) = \cos(\mathbf{e}_{c_s}, \mathbf{e}_{c_t}),
\]
and partition motifs into ten uniform bins over $[0,1]$ with width 0.1. Bins are ordered in descending similarity, prioritizing semantically coherent edits. Within each bin, motifs are further ranked by their occurrence frequency $w_{\mathcal{T}}(m)$, which captures empirical priors over affective rule selection for a given source concept.

\noindent \textbf{Overview of The Constructed EmoTree.}
Applying the above procedure to all images in EmoSet and both target valence directions yields a large repository of structured affective rules. For each semantic cluster and source concept, the EmoTree $\mathcal{T}$ stores a set of candidate target motifs organized according to emotional direction. Overall, $\mathcal{T}$ contains 10,848 positive and 13,505 negative target concepts connected through more than 320,000 target attributes, actions, and scene contexts. The resulting structure forms an interpretable lexicon of compositional visual motifs grounded in large-scale image--emotion data. A partial visualization of $\mathcal{T}$ is shown in \textbf{Fig.~\ref{fig:fig4}A}.

\noindent \textbf{Hierarchical EmoTree Search during Inference.}
Given an input image $I$ and a target valence $y_t$, EROS traverses $\mathcal{T}$ hierarchically to retrieve an appropriate affective rule.
First, the emotional valence of the image is estimated using the recognition module (Module 1), and a CLIP image embedding is extracted: $\mathbf{v}_I = f_{\mathrm{CLIP}}^{\mathrm{img}}(I)$. The most similar semantic cluster is then retrieved via
\[
k^* = \arg\max_k \cos(\mathbf{v}_I, \bar{\mathbf{v}}_k).
\]
where $\bar{\mathbf{v}}_k$ denotes the prototype embedding of cluster $C_k$.

Within the selected cluster $C_{k^*}$, we generate an image caption $t$ using BLIP and extract a source concept candidate $c_s^{\mathrm{query}}$ using spaCy~\cite{honnibal2020spacy} by selecting the primary noun phrase (or the full caption when no noun phrase is identified). Unlike EmoTree construction, which uses GPT-4o for offline concept extraction, inference relies exclusively on local models. This design removes dependence on proprietary large multimodal models during deployment and enables privacy-preserving operation on personal devices.

The extracted concept is then matched to the closest source concept stored in the selected cluster:
\[
c_s^* = \arg\max_{c_s \in \mathcal{S}(C_{k^*})} \cos(\mathbf{e}_{c_s^*}, \mathbf{e}_{c_s}),
\]
where $\mathbf{e}_{c_s} = f_{\mathrm{CLIP}}^{\mathrm{text}}(c_s)$. Given the retrieved source concept $c_s^*$ and the desired target valence $y_t$, EROS traverses the corresponding branch of $\mathcal{T}$ to retrieve a ranked set of candidate target motifs. These motifs serve as affective intervention plans that guide the subsequent image-generation module.

Candidate motifs are ranked according to their similarity bin and occurrence frequency. By default, EROS prioritizes motifs from the highest-similarity bin and selects the motif with the highest frequency within that bin, thereby favoring semantically coherent edits that minimally alter the source image.
However, this strategy may be suboptimal for cross-valence transformations involving strongly valenced source concepts. In such cases, highly similar target concepts often preserve the semantic attributes responsible for the original emotional response, thereby limiting effective affective transfer. For example, in a negatively valenced image depicting a burning house, replacing the concept fire with semantically similar concepts such as smoke or flames is unlikely to alter the scene's emotional interpretation, as these concepts remain strongly associated with danger and destruction. Achieving a positive emotional transformation may instead require replacing fire with a semantically distant concept, such as flowers or rainbow decorations, which fundamentally changes the semantic context and consequently the affective meaning of the scene.

To encourage meaningful emotional transformations, we modify the search strategy for cross-valence editing. Specifically, when the retrieved source concept is not neutral ($c_s^* \notin \mathcal{S}_{\mathrm{neu}}$) and the source and target valences differ ($y_s \neq y_t$), motifs with source--target concept similarity greater than 0.8 are skipped during retrieval. Search instead begins from the highest-ranked remaining similarity bin. In all other cases, retrieval proceeds from the highest-similarity bin as described above.

The top-ranked motif is selected as the affective intervention plan unless it is rejected by the personalization module (Module 5). In that case, retrieval continues to the next candidate motif. This hierarchical search procedure balances semantic preservation with emotional controllability, enabling EROS to retrieve editing rules that are both contextually coherent and affectively effective.

\subsubsection{Module 4 -- Generation: Synthesizing Images from Controlled Prompts}
\label{sec:module4_gen}

The generation module executes the affective intervention proposed by the EmoTree to produce edited images through localized, structure-preserving modifications (\textbf{Fig.~\ref{fig:S5}}). Given an input image $I$ and a target motif $m=(c_t,a_t,r_t,s_t)$ retrieved via hierarchical EmoTree search (Module~3, \textbf{Sec.\ref{sec:module3_EmoTree}}), we first convert the motif into a textual prompt $p$ using template-based composition.

Specifically, we define a template set $\mathcal{Q}$ that maps $(c_t,a_t,r_t,s_t)$ into natural-language descriptions. Each template integrates the target concept, attribute, action, and scene context into a structured sentence. For example, a template may take the form ``The $\{a_t\}$ $\{c_t\}$ $\{r_t\}$ $\{s_t\}$'', where the placeholders are instantiated using the selected motif. To maintain grammatical correctness, single-word attributes are used as direct modifiers of the target concept (e.g., ``the ${a_t}$ ${c_t}$''), whereas multi-word attributes are expressed as descriptive clauses (e.g., ``with ${a_t}$''). A template $Q\sim\mathcal{Q}$ is sampled to generate
$p = Q(c_t, a_t, r_t, s_t)$, followed by light linguistic normalization, including preposition insertion and word-order adjustment for scene phrases. Examples of generated prompts are shown in \textbf{Fig.~\ref{fig:fig4}F}.

Image editing is performed through an iterative inpainting procedure that combines the prompt $p$ with spatial constraints defined by the binary mask $M_{\text{final}}\in[0,1]^{H\times W}$ obtained from Module~2 (\textbf{Sec.\ref{sec:module2_region}}). 
For same-valence editing, where the valence predicted by $\mathcal{P}$ for the source image matches the target valence, we preserve the localized affective region and modify the surrounding context by inverting the mask, $M_{\text{edit}} = 1 - M_{\text{final}}$.
For cross-valence editing, the affective region itself is modified, $M_{\text{edit}} = M_{\text{final}}$. 

This strategy reflects two distinct intervention mechanisms. Same-valence editing strengthens existing affective cues by enriching the surrounding context while preserving the primary emotion-carrying subject. In contrast, cross-valence editing requires direct modification of the visual elements responsible for the original emotional response in order to induce a change in valence. For example, same-valence enhancement may be achieved by enriching the surrounding context of a smiling boy with positively associated elements such as balloons or fireworks while preserving the subject itself. In contrast, cross-valence editing generally requires modifying the emotion-carrying subject (e.g., facial expression) to alter the perceived emotional valence.

We perform conditional image generation using a Stable Diffusion inpainting model~\cite{rombach2022SD}. The model receives the original image $I$, editing mask $M_{\text{edit}}$, structural prior $R$, and prompt $p$ as inputs. The structural prior $R = f_{\text{ctrl}}(I)$, is obtained using ControlNet~\cite{zhang2023controlnet} and encodes edge-based structural information that preserves the global layout and geometry of the scene during generation. The edited image is generated as
\[
\tilde{I} = \mathcal{G}(I, M_{\text{edit}}, R, p),
\]
where $\mathcal{G}$ denotes the conditional diffusion process.

During generation, we use UniPC scheduling and DDIM sampling (50 steps, guidance scale 7.5, ControlNet scale 1.0). A negative prompt (``low quality, blurry, distorted'') is included to suppress common generation artifacts. All computations are performed using FP16 precision.

Although Stable Diffusion inpainting is used throughout our experiments, EROS is agnostic to the underlying text-conditioned image generator. The compositional prompts derived from the EmoTree can be integrated with a variety of image-editing models. We demonstrate this by applying the same prompts across SD-Inpainting~\cite{rombach2022SD}, Nano Banana~\cite{nanobanana}, and GPT-Image1~\cite{GPT-Image1} 
(\textbf{Fig.~\ref{fig:S8}}).
Differences in visual quality therefore primarily reflect the capabilities of the underlying generative model rather than the affective reasoning process itself.

To ensure effective affective transformation, we adopt an iterative refinement procedure. At each iteration, the generated image $\tilde{I}$ is evaluated using the emotion predictor $\mathcal{P}$ (Module~1, \textbf{Sec.\ref{sec:module1_recog}}), yielding a confidence score $p(y_t \mid \tilde{I})$. Generation terminates when this score exceeds a threshold $\gamma = 0.7$ or when the minimum spatial extent constraint is reached ($\tau = 0$; Module~2,  \textbf{Sec.~\ref{sec:module2_region}}). 
If neither condition is satisfied, the editing mask is progressively expanded by reducing $\tau$, thereby increasing the editable region in subsequent iterations. This coarse-to-fine refinement strategy prioritizes localized modifications and expands the intervention only when necessary, balancing emotional transformation with preservation of scene structure and semantic content.

\subsubsection{Module 5 -- Personalization: Adapting to User Preferences}
\label{sec:module5_Personalization}

The personalization module captures individual emotional preferences and sensitivities by maintaining an interpretable user-specific affective profile. This enables inference-time adaptation, allowing EROS to personalize affective interventions without additional training or fine-tuning.

While the EmoTree $\mathcal{T}$ defines a global repository of affective editing rules, emotional responses vary substantially across individuals. To account for this variability, we introduce a user-specific memory bank, \textit{EmoMem} ($\mathcal{H}$), which records feedback from previous interactions. Following each edited image, users indicate whether the modification successfully evokes the target valence. EROS then updates EmoMem and adapts future motif selection based on the accumulated feedback.

\noindent \textbf{EmoMem ($\mathcal{H}$).}
For each target valence, EmoMem consists of two complementary memory banks: an accepted-memory bank $\mathcal{H}^{acc}$ and a rejected-memory bank $\mathcal{H}^{rej}$. 
$\mathcal{H}^{acc}$ stores motifs from edits that users accepted as successfully evoking the target emotional valence, whereas $\mathcal{H}^{rej}$ stores motifs from edits that users rejected because they did not evoke the intended emotion. The distinction between the two memory banks is therefore entirely determined by each user's individual affective judgment. For example, one user may perceive a \textit{skull} as exciting or aesthetically appealing and store it in $\mathcal{H}^{acc}$, whereas another may find the same motif disturbing and instead store it in $\mathcal{H}^{rej}$. Together, these complementary memories enable EROS to learn personalized associations between visual motifs and emotional responses, supporting future affective editing through reusable symbolic memories.

Without loss of generality, we describe the construction of the positive-valence memories, $\mathcal{H}^{pos,acc}$ and $\mathcal{H}^{pos,rej}$. For simplicity, we omit the superscript ``pos'' in the following description. An identical pair of accepted and rejected memory banks is maintained for the negative valence. Importantly, accepted and rejected motifs are conditioned on the same target valence and therefore do not differ in emotional polarity. Instead, they encode relative user preference among multiple valid affective interventions.

The accepted-memory bank $\mathcal{H}^{acc}$ stores visual motifs that have been validated to successfully evoke the target emotional response for a particular user.
Each memory entry is represented as the tuple
$
(\mathbf{v}, y_t, m),
$
where $\mathbf{v}$ denotes the embedding of the source image $I$, $y_t$ is the target valence, and $m$ is the corresponding visual motif. By jointly storing source-image representations and validated motifs, $\mathcal{H}^{acc}$ enables EROS to retrieve previously preferred affective interventions when encountering visually similar images during future interactions. Consequently, personalization extends beyond individual symbols to reusable affective editing strategies.

The rejected-memory bank $\mathcal{H}^{rej}$ stores visual motifs that have been judged not to evoke the target emotional response for the same user.
Rather than storing rejected motifs as indivisible units, each rejected motif
$
m=(c_t,a_t,r_t,s_t)
$
is decomposed into its constituent symbols and recorded in four separate rejection sets: $\mathcal{H}_{con}^{rej}$ for concepts, $\mathcal{H}_{att}^{rej}$ for attributes, $\mathcal{H}_{act}^{rej}$ for actions, and $\mathcal{H}_{sce}^{rej}$ for scene contexts. Each symbol maintains an associated rejection count that records how frequently it has been rejected.

Unlike $\mathcal{H}^{acc}$, the rejected-memory bank does not store source-image information. The objective is not to avoid specific image--motif pairings, but rather to suppress individual symbols that consistently fail to produce the desired emotional response, regardless of image context. This decomposition enables fine-grained adaptation and prevents repeatedly proposing concepts, attributes, actions, or scenes that a user systematically dislikes.

To further capture strong preferences, we maintain a hard-rejection set $\mathcal{H}_{rej}^{hard}$. Concepts that are rejected in more than three consecutive interactions are added to this set and are subsequently excluded from future motif generation. The hard-rejection mechanism provides a simple yet effective means of modeling persistent aversions while preserving the flexibility of the broader memory system.

The asymmetric design of $\mathcal{H}^{acc}$ and $\mathcal{H}^{rej}$ reflects their distinct functional roles. Accepted memories store reusable affective intervention strategies that can be transferred across visually similar images, whereas rejected memories operate at the symbol level to suppress undesirable concepts and encourage exploration of alternative affective motifs.

\noindent \textbf{Personalized Affective Rule Selection.}
At inference time, given an input image $I$ and target valence $y_t$, Module~3 (\textbf{Sec.\ref{sec:module3_EmoTree}}) retrieves a candidate motif $m^{\mathcal{T}}_{\text{cand}}$ through hierarchical search over the EmoTree $\mathcal{T}$.
In parallel, EROS retrieves a personalized candidate from the accepted-memory bank $\mathcal{H}^{acc}$ from the target valence. Let
$
\mathbf{v}I=f_{\mathrm{CLIP}}^{\mathrm{img}}(I)
$
denote the CLIP embedding of the input image. For each stored tuple $(\mathbf{v}, y_t, m)$ in $\mathcal{H}^{acc}$, we compute the cosine similarity between $\mathbf{v}$ and $\mathbf{v}_I$. Among entries with matching target valence $y_t$, the motif associated with the most similar image is selected, yielding a personalized candidate motif $m^{\mathcal{H}}_{\text{cand}}$.
The two candidates serve complementary roles. The EmoTree candidate $m^{\mathcal{T}}_{\text{cand}}$ is derived from globally learned affective rules, whereas the memory-based candidate $m^{\mathcal{H}}_{\text{cand}}$ corresponds to a motif that has previously been validated by the user. Both candidates are subsequently filtered using user-specific rejection memories.

Without loss of generality, we describe the filtering procedure for $m^{\mathcal{T}}_{\text{cand}}$; the same procedure is applied to $m^{\mathcal{H}}_{\text{cand}}$. Given a candidate motif
$
m=(c_t,a_t,r_t,s_t),
$
we first evaluate its target concept $c_t$. If
$
c_t \in \mathcal{H}_{con}^{hard},
$
the motif is immediately discarded and the next candidate is considered. Otherwise, if
$
c_t \in \mathcal{H}_{con}^{rej},
$
it is probabilistically rejected according to
$
p^{\text{rej}}(c_t)=
\frac{N(c_t)}
{\sum_{c' \in \mathcal{H}_{con}^{rej}} N(c')},
$
where $N(\cdot)$ denotes the rejection count associated with the concept. If the concept is rejected, the entire motif is discarded. Otherwise, the remaining motif components $a_t$, $r_t$, and $s_t$ are evaluated independently using their corresponding rejection memories $\mathcal{H}{att}^{rej}$, $\mathcal{H}{act}^{rej}$, and $\mathcal{H}_{sce}^{rej}$ with the same frequency-based rejection rule. Rejected components are removed, whereas accepted components are retained.
If both $m^{\mathcal{T}}{\text{cand}}$ and $m^{\mathcal{H}}{\text{cand}}$ fail the rejection criteria, retrieval proceeds to the next candidate from each source. If one source exhausts all available candidates, the highest-ranked valid motif from the remaining source is selected.

When both candidates survive rejection filtering, the final selection is determined according to visual similarity to the input image. For the EmoTree candidate $m^{\mathcal{T}}_{\text{cand}}$, cosine similarity is measured using the prototype embedding $\bar{\mathbf{v}}_k$ of the associated semantic cluster $C{k^*}$, yielding
$
\cos(\mathbf{v}_I,\bar{\mathbf{v}}_k).
$
For the memory-based candidate $m^{\mathcal{H}}_{\text{cand}}$, cosine similarity is computed as
$
\cos(\mathbf{v}_I,\mathbf{v}),
$
where $\mathbf{v}$ is the stored image embedding associated with the retrieved memory entry. The motif with the higher similarity score is selected as the final motif
$
m^*=(c_t,a_t,r_t,s_t).
$

The selected motif $m^*$ is then passed to Module~4 (\textbf{Sec.\ref{sec:module4_gen}}) for controlled image editing. Following image generation, binary user feedback is collected. If the edited image is accepted, the corresponding tuple $(\mathbf{v},y_t,m^*)$ is added to $\mathcal{H}^{acc}$. Otherwise, the motif is decomposed into its constituent symbols and the corresponding rejection memories are updated. Through this human-in-the-loop process, EmoMem $\mathcal{H}$ is continuously refined, enabling personalized affective editing to adapt progressively to individual user preferences over time.

\subsection{Model Variants of EROS}
\label{sec:EROS_Variants}

The default EROS framework combines two key mechanisms: (i) hierarchical retrieval of affective rules from the EmoTree $\mathcal{T}$ and (ii) inference-time personalization through the user-specific memory bank $\mathcal{H}$. To quantify the contribution of each component, we construct the following model variants by selectively disabling individual mechanisms.

\noindent \textbf{EROS-1loop.}
This variant performs hierarchical search over $\mathcal{T}$ only once, generating a single edited output for each input image. No iterative motif retrieval or refinement is performed. This setting evaluates whether a single affective intervention derived from the most semantically aligned and frequently occurring rule is sufficient for effective affective image editing, and quantifies the quality of the affective rules encoded in $\mathcal{T}$ independent of iterative motif retrieval.

\noindent \textbf{EROS w/o Personalization Feedback (EROS-NoMem).}
This variant disables inference-time personalization by preventing $\mathcal{H}$ from participating in motif selection. All affective interventions are therefore derived exclusively from hierarchical search over $\mathcal{T}$.
To ensure a controlled comparison, the memory bank $\mathcal{H}$ continues to be updated using user feedback throughout the interaction process. However, the stored memories are never consulted during retrieval or motif selection. Consequently, the model retains the same editing and feedback pipeline as the full EROS framework while removing the influence of personalized affective memories.
This setting isolates the contribution of the user-specific memory bank and evaluates the extent to which personalization improves affective image editing beyond the global affective rules encoded in $\mathcal{T}$.

\subsection{Baselines}
\label{sec:baselines}

We compare EROS against several representative baselines. Methods capable of incorporating user feedback are evaluated in the interactive personalized image-editing setting. Methods that do not support iterative adaptation are evaluated only during the first interaction round, without human feedback in subsequent loops.

\subsubsection{Large Model Series (LMS) and Variants}

Large multimodal models (LMMs) have demonstrated strong capabilities in image understanding, reasoning, and in-context learning~\cite{hurst2024gpt,openai2025o4}. To evaluate whether these capabilities are sufficient for personalized affective image editing, we construct a baseline termed the \textit{Large Model Series} (LMS), which combines multiple LMMs within an iterative editing framework.

At each loop, GPT-4o~\cite{hurst2024gpt} first generates a caption of the input image using the prompt: \textit{``Describe the image content in one sentence without any introductory phrases like `The image shows'. Start immediately with the scene details.''} The resulting caption, target valence, and user preference memory are then provided to GPT-o4~\cite{openai2025o4}, which generates an editing instruction using the prompt: \textit{``The original image depicts: ${t_i}$. The desired emotional valence is: ${y_t}$. User likes the following concepts: ${\mathcal{C}^{acc}_{\text{LMS}}}$. User dislikes the following concepts: ${\mathcal{C}^{rej}_{\text{LMS}}}$. Generate a new description (less than 15 words) that modifies the image to clearly express this emotion.''}
The generated editing instruction is subsequently passed to a diffusion model for image synthesis. Although GPT-4o supports native image generation, we instead employ the same diffusion backbone~\cite{rombach2022SD} as EROS (\textbf{Sec.\ref{sec:EROS}}) to isolate affective reasoning and personalization from differences in image-generation capability.

At the first loop, the user-specific memory is empty. In subsequent loops, edited images are evaluated by the user and re-captioned using the same GPT-4o captioning prompt. Concepts are extracted from the resulting descriptions using spaCy~\cite{honnibal2020spacy} by identifying the primary noun phrase. Based on user feedback, extracted concepts are added to either the accepted-memory set ${\mathcal{C}^{acc}_{\text{LMS}}}$ or the rejected-memory set ${\mathcal{C}^{rej}_{\text{LMS}}}$ of the corresponding target valence, which are subsequently incorporated into future editing prompts.

We further define two LMS variants that mirror the EROS ablations. Both variants use the same captioning, instruction-generation, and image-synthesis pipeline as the default LMS. They differ only in whether iterative refinement and personalization are enabled.

\noindent \textbf{LMS-1Loop.}
This variant generates a single edited output for each input image during the first interaction round. No user feedback is collected and no subsequent editing loops are performed. This setting evaluates whether a single LMM-generated affective intervention is sufficient for effective affective image editing.

\noindent \textbf{LMS w/o Personalization Feedback (LMS-NoMem).}
This variant disables the use of the accepted-memory and rejected-memory sets, $\mathcal{C}^{acc}_{\text{LMS}}$ and $\mathcal{C}^{rej}_{\text{LMS}}$, during instruction generation. Iterative editing is still performed, but each editing round is conditioned only on the current image and target valence, without access to accumulated user preferences.
For a controlled comparison, $\mathcal{C}^{acc}_{\text{LMS}}$ and $\mathcal{C}^{rej}_{\text{LMS}}$ continue to be updated for each valance using user feedback throughout the interaction process. However, the stored memories are never used to guide subsequent edits. This setting isolates the contribution of memory-based personalization within the LMS framework.

\subsubsection{XDream and its Variants}

The original XDream framework~\cite{ponce2019XDream} evolves visual stimuli through evolutionary optimization in latent space to maximize neural responses in biological brains. We adapt XDream to the personalized affective image-editing setting by replacing its original generator with InstructPix2Pix (Ip2p)~\cite{brooks2023instructpix2pix}, enabling direct editing of a source image conditioned on text embeddings.
Instead of initializing the search with random latent codes, we initialize the embedding using CLIP~\cite{radford2021CLIP} text features derived from a generic affective instruction template (``To $<$target valence$>$''), with additive random perturbations to encourage diversity. During optimization, the embedding is iteratively mutated, and each updated embedding is used to guide Ip2p to generate a candidate edited image.

Unlike EROS and LMS, XDream does not perform symbolic affective reasoning and does not maintain an explicit memory of accepted or rejected user preferences. Personalization emerges solely through evolutionary search driven by human feedback, making XDream a representative feedback-driven optimization baseline.

\noindent \textbf{XDream-1Loop.}
This variant disables iterative mutation and returns only the first edited image generated from the initial text embedding. It evaluates the contribution of evolutionary refinement within the XDream framework.

\subsubsection{Real Images (Real)}
Because EmoSet~\cite{yang2023emoset} contains real-world images annotated with emotional valence, we include a retrieval-based baseline that directly returns images from EmoSet as affective exemplars. This baseline serves as a control condition that prioritizes emotional alignment without preserving structural correspondence to the source image, thereby highlighting the trade-off between emotional effectiveness and source-image fidelity.
Unlike EROS, LMS, and XDream, this setting does not perform image editing. Instead, for a given target valence, images are randomly sampled from EmoSet and returned as outputs. At each feedback loop, a new image is sampled independently.

\subsubsection{Non-Interactive Image Editing Methods}

Most existing affective image-editing methods neither support personalization nor allow iterative refinement. We therefore compare them against the single-step variants of our interactive methods (EROS-1Loop, LMS-1Loop, XDream-1Loop, and Real). Each baseline generates a single edited image per trial.

\noindent \textbf{(1) Color Transfer (CT)}~\cite{pitie2007automated} matches the global color distribution of a source image to that of a reference image sampled from EmoSet according to the target valence.

\noindent \textbf{(2) Neural Style Transfer (NST)}~\cite{gatys2015neural} transfers style through deep feature statistics using reference images sampled from EmoSet according to the target valence.

\noindent \textbf{(3) CLIP-Styler (CSty)}~\cite{kwon2022clipstyler} performs CLIP-guided text-driven style transfer conditioned on a one-word target-valence prompt.

\noindent \textbf{(4) SDEdit}~\cite{meng2021sdedit} performs diffusion-based image editing conditioned on the source image and a one-word target-valence prompt.

\noindent \textbf{(5) ControlNet (CtrlNet)}~\cite{zhang2023controlnet} performs structure-preserving diffusion-based editing conditioned on the source image and a one-word target-valence prompt.

\noindent \textbf{(6) InstructPix2Pix (Ip2p)}~\cite{brooks2023instructpix2pix} performs instruction-based image editing using a conditional diffusion model, conditioned on the source image and a one-word target-valence prompt.

\noindent \textbf{(7) BLIP-Diffusion (BlipDiff)}~\cite{li2023blip} performs subject-driven image editing using a vision--language encoder conditioned on the source image and a one-word target-valence prompt.

\noindent \textbf{(8) Affective Image Filter (AIF)}~\cite{weng2023affective} performs emotion-guided global image transformations conditioned on the source image and a one-word target-valence prompt.

\subsubsection{Implementation Details}

All experiments were conducted on NVIDIA RTX A6000 GPUs or through API access to proprietary LMMs. EROS was implemented in PyTorch and optimized using Adam. For all baselines, we used the official implementations and publicly released checkpoints whenever available. Style-transfer methods (CT, NST, and CSty) were evaluated using their original configurations. Diffusion-based methods (SDEdit, CtrlNet, Ip2p, BlipDiff, and AIF) were evaluated using their default inference settings. For CT and NST, reference images were randomly sampled from EmoSet according to the target valence.

\subsection{Human Psychophysics Experiments}
\label{sec:exp_all}
To comprehensively evaluate the emotional intelligence of AI systems, including EROS, we conducted a series of human psychophysics experiments targeting four core capabilities: emotion reasoning (identifying the causes of emotion), emotion control (deriving affective intervention strategies), emotion-guided generation (producing images that reliably evoke a desired emotional response), and personalization (capturing user-specific affective preferences). Unless otherwise stated, all source images were drawn from the EmoSet test set. Each participant took part in the study only once. All procedures were conducted with informed consent under protocols approved by the Institutional Review Board of our institution, and participants received appropriate compensation.

\subsubsection{Exp-EmoInteract: Personalized Interactive Affective Editing}
\label{sec:Exp-EmoInteract}

To evaluate personalized affective image editing in an interactive human-in-the-loop setting, we conducted an in-lab psychophysics experiment (\textbf{Fig.~\ref{fig:fig2}B}). 

\noindent \textbf{Experimental Protocol.}
In each trial, participants were presented with a source image, a target valence, and edited outputs generated by EROS and LMS. The two outputs were shown sequentially in randomized order. For each output, participants independently judged whether the edited image successfully evoked the target valence.
Based on these responses, three outcomes were possible. (1) If both methods successfully evoked the target valence, participants indicated which edited image better preserved the structural and semantic content of the source image. (2) If only one method succeeded, the unsuccessful method proceeded to the next editing loop while the successful method terminated. (3) If neither method succeeded, both methods proceeded to the next loop. This procedure continued until a successful edit was obtained or the maximum number of loops was reached.

\noindent \textbf{Experimental Details.}
We recruited 41 participants, each evaluating 20 source images. For every image, participants assessed both methods independently. The order of images and the presentation order of methods were randomized across trials.
To ensure participant attention and data quality, we interleaved 6 control trials throughout the experiment. In each control trial (\textbf{Fig.\ref{fig:S9}A}), participants were shown a source image together with two candidate images presented sequentially in randomized order: (i) an identical copy of the source image and (ii) an edited image that clearly expressed the target valence. Participants judged whether each candidate successfully evoked the target valence relative to the source image. A valid response required labeling the source-identical image as ``No'' and the emotionally edited image as ``Yes''. Because each image was evaluated independently, this yielded 12 control decisions per participant.
Participants with more than 25\% incorrect control responses ($>$3/12) were excluded from further analysis. The control set contained three positive-valence and three negative-valence trials. After quality-control filtering, 35 participants were retained for analysis.

\noindent \textbf{Evaluation Criteria.}
This experiment evaluates four aspects of interactive affective image editing.
\textit{Validity} measures whether an edited image successfully evokes the target valence at the first interaction round. 
\textit{Efficiency} measures how rapidly a method reaches a successful edit, quantified by the loop at which success first occurs.
\textit{Preference} measures participant preference when competing methods successfully evoke the target valence within the same interaction round.
\textit{Fidelity} measures the extent to which the edited image preserves the structural and semantic content of the source image while introducing the desired emotional modification.
These criteria correspond to complementary dimensions of emotional intelligence, namely emotional effectiveness, adaptation efficiency, user preference, and content preservation. We report model performance using success rate, cumulative success rate, preference rate, SSIM-C, and L1-C (\textbf{Sec.\ref{sec:metric}}), with detailed results presented in \textbf{Sec.\ref{sec:in-lab}}.

\subsubsection{Exp-EmoPairwise: Comparison with State-of-the-Art Image Editing Methods}
\label{sec:Exp-EmoPairwise}

To evaluate the effectiveness of EROS, we conducted a large-scale human preference experiment (Exp-EmoPairwise) comparing our method against a diverse set of affective image-editing approaches (\textbf{Fig.~\ref{fig:fig1}D}). Because most existing methods do not support personalization or iterative refinement, we compared them against the single-step variants EROS-1Loop, LMS-1Loop, XDream-1Loop, and Real (\textbf{Sec.~\ref{sec:baselines}}). All methods produced a single edited image per trial.

\noindent \textbf{Experimental Protocol.}
In each trial, participants were presented with a source image and two edited images: one generated by EROS-1Loop and the other generated by a competing baseline (\textbf{Sec.~\ref{sec:baselines}}). Participants performed a binary forced-choice task, selecting the image that better conveyed the target emotional valence while remaining visually consistent with the source image.
The instruction shown to participants was: \textit{``Consider the reference image shown above. Which edited image better evokes the target emotion?''} (\textbf{Fig.\ref{fig:fig1}D}). Although structural fidelity was not explicitly mentioned, the presence of the reference image encouraged participants to jointly evaluate emotional effectiveness and preservation of source-image content.

\noindent \textbf{Experimental Details.}
We recruited 79 participants through Amazon Mechanical Turk (MTurk)~\cite{turk2012amazon}. Each participant evaluated 15 source images. For every image, EROS-1Loop was compared pairwise against the outputs of 11 baseline methods, resulting in 165 comparison trials per participant. The order of source images and the presentation order of the two candidate images were randomized independently for each participant.

To ensure participant attention and data quality, six control trials were interleaved with the main experiment (two target valences, three trials per valence). In each control trial (\textbf{Fig.\ref{fig:S9}B}), participants completed the same binary preference task. Control image pairs were intentionally constructed to be unambiguous, with one image clearly expressing the target valence and the other expressing the opposite valence. Trial order and image presentation were randomized.
Participants with more than one incorrect response on the control trials ($>$1/6) were excluded from further analysis. After quality-control filtering, data from 76 participants were retained.

\noindent \textbf{Evaluation Criteria.}
This experiment evaluates \textit{Preference}, which measures the relative ability of competing methods to generate emotionally effective edits under a single editing step. Preference is quantified as the proportion of trials in which a method is selected over its paired competitor. We report preference rate (\textbf{Sec.\ref{sec:metric}}) in the results (\textbf{Sec.\ref{sec:pairwise}}).

\subsubsection{Exp-EmoRegion: Emotion-Relevant Region Annotation}
\label{sec:Exp-EmoRegion}

To identify image regions that humans perceive as responsible for emotional responses, we conducted a human annotation experiment (Exp-EmoRegion) in which participants localized the visual regions they considered most influential for the perceived affect of an image (\textbf{Fig.~\ref{fig:fig3}B1}).

\noindent \textbf{Experimental Protocol.} In each trial, participants were presented with a source image and a target emotional valence. Using a computer mouse, they manually delineated the image regions that they believed contributed most strongly to the specified emotional response. Annotations were collected as binary region masks.

\noindent \textbf{Experimental Details.} Annotations were collected from 46 participants on Amazon Mechanical Turk (MTurk)~\cite{turk2012amazon}. Each participant annotated 50 source images, yielding a total of 2,300 annotation trials. Images were sampled from a pool of 500 source images, with each image annotated by at least three independent participants. Images were assigned randomly to annotators, and presentation order was randomized to minimize potential order effects. The same attention-control trials used in Exp-EmoPairwise (\textbf{Sec.~\ref{sec:Exp-EmoPairwise}}) were interleaved throughout the experiment. Participants who failed the control criteria were excluded from further analysis. After quality-control filtering, data from 46 participants were retained.

\noindent \textbf{Evaluation Criteria.} Binary masks from different annotators were retained as independent human annotations. These annotations support two complementary evaluations. First, they provide a measure of within-human consistency in identifying emotion-relevant regions. Second, they enable human--EROS comparisons (\textbf{Fig.~\ref{fig:fig3}B2}) to assess whether the regions localized by EROS align with human judgments. We quantify agreement using the Intersection-over-Union (IoU) metric (\textbf{Sec.~\ref{sec:metric}}) and report both human--human and human--EROS consistency in the results (\textbf{Sec.~\ref{sec:res_EmoRegion}}).

\subsubsection{Exp-EmoPrompt: Affective Rule Evaluation}
\label{sec:Exp-EmoPrompt}

To evaluate whether the EmoTree $\mathcal{T}$ captures meaningful affective knowledge, we conducted a human psychophysics experiment (Exp-EmoPrompt) assessing whether textual descriptions derived from EmoTree motifs convey the intended emotional valence (\textbf{Fig.~\ref{fig:fig5}A1}). Each prompt corresponds to a natural-language description $p$ generated from a motif (\textbf{Sec.\ref{sec:module4_gen}}), collectively specifying the concept, attributes, actions, and scene context associated with an affective rule.

\noindent \textbf{Experimental Protocol.} In each trial, participants were presented with a motif description $p$ and asked to determine whether it conveyed a positive or negative emotional valence. The presentation order of prompts and the response options (``positive'' and ``negative'') were randomized to minimize response bias.

\noindent \textbf{Experimental Details.} We recruited 79 participants through Amazon Mechanical Turk (MTurk)~\cite{turk2012amazon}. Each participant evaluated 100 motif descriptions generated by EROS, comprising 50 positive-valence and 50 negative-valence prompts. Six control trials (three positive and three negative) containing unambiguous emotional descriptions were interleaved throughout the experiment as attention checks (\textbf{Fig.\ref{fig:S9}C}). Responses were considered incorrect when the participant-selected label did not match the intended valence associated with the motif. Participants who failed more than one control trial ($>$1/6) were excluded from further analysis. After quality-control filtering, data from 44 participants were retained.

\noindent \textbf{Evaluation Criteria.} This experiment evaluates the extent to which affective rules encoded in the EmoTree align with human emotional judgments. High agreement indicates that the compositional motifs learned by EROS capture affective knowledge that is both interpretable and consistent with human perception. We quantify this agreement using the alignment score (\textbf{Sec.\ref{sec:metric}}) and report the results in \textbf{Sec.\ref{sec:res_emoprompt}}.

\subsubsection{Exp-EmoRule: Evaluation on Generic and Personalized Affective Rules}
\label{sec:Exp-EmoRule}

To evaluate whether EROS learns effective generic affective rules in the EmoTree $\mathcal{T}$ and accurate personalized affective profiles in EmoMem $\mathcal{H}$, we designed a three-stage human psychophysics experiment. For comparison, we included three baselines (LMS-NoMem, XDream, and Real; \textbf{Sec.\ref{sec:baselines}}). None of these methods employ an explicit user-specific memory bank to guide future edits, although they track user responses across loops.

\paragraph{Stage 1: Evaluation of Generic Affective Rules Through Image Editing.}
This stage evaluates whether the affective rules encoded in $\mathcal{T}$ produce stronger emotional responses and more reliably achieve the target valence than competing approaches (\textbf{Fig.~\ref{fig:fig5}B1}). To isolate the contribution of $\mathcal{T}$, we use EROS-NoMem and present edited images generated from the top five retrieved motifs sequentially across five loops.

\noindent \textbf{Experimental Protocol.} In each trial, participants were shown a source image, a target valence, and two edited images generated by EROS-NoMem and a competing baseline at the same loop, together with a ``none of them'' option. Participants selected the image that best matched the target valence or rejected both images. If one method was selected, the unselected image was subsequently compared against ``none of them'' to determine whether it weakly expressed the target valence or failed entirely. If neither method was selected, both methods proceeded to the next loop. The process continued until both methods achieved the target valence or the maximum of five loops was reached.

\paragraph{Stage 2: Precision of Personalized Affective Profiles.}
This stage evaluates how accurately different methods capture user-specific affective preferences. Concepts associated with successful edits in Stage 1 were stored in a user-specific memory for each method. We then directly evaluated the quality of the resulting personalized concept representations (\textbf{Fig.~\ref{fig:fig6}A1}).

\noindent \textbf{Experimental Protocol.} In each trial, participants were presented with three concept sets corresponding to EROS-NoMem, LMS, and XDream. Each set contained four concepts sampled from the method-specific memory conditioned on the target valence. Participants selected the concept set that best matched their personal preferences. Two independent evaluations were conducted for each valence using non-overlapping concept sets to reduce repetition effects. When fewer than four concepts were available in a method's memory, the remaining concepts were randomly sampled from a method-specific global concept pool. 
For EROS-NoMem, the pool consisted of all target concepts $c_t$ retrieved from EmoTree motifs. For LMS and XDream, concepts in the pool were extracted from captions generated from edited images using GPT-4o~\cite{hurst2024gpt} and BLIP~\cite{li2022blip}, respectively, followed by noun-phrase extraction using spaCy~\cite{honnibal2020spacy}.

\paragraph{Stage 3: Consistency of Personalized Affective Profiles.}

This stage evaluates whether the personalized affective profile learned by EROS reflects stable and internally consistent preferences rather than incidental feedback accumulated during interaction (\textbf{Fig.~\ref{fig:fig6}B1}). If EmoMem $\mathcal{H}$ accurately captures user preferences, concepts stored in the accepted memory should consistently align with a participant's target valence, whereas concepts stored in the rejected memory should systematically diverge from it.

\noindent \textbf{Experimental Protocol.}
Using the accepted ($\mathcal{H}_{con}^{acc}$) and rejected ($\mathcal{H}_{con}^{rej}$) concept memories in EROS (\textbf{Sec.\ref{sec:module5_Personalization}}), we constructed three types of concept sets conditioned on the target valence: all-accepted, half-accepted, and all-rejected. Participants performed pairwise comparisons between these sets and selected the set that best matched the target valence. Two independent rounds were conducted for each valence using non-overlapping concept samples.\\

\noindent \textbf{Experimental Details in All Three Stages.}
We recruited 51 participants through Amazon Mechanical Turk (MTurk)~\cite{turk2012amazon}. Each participant completed all three stages. Trial order within each stage was randomized, whereas the order of stages was fixed. Attention-control trials were included in all stages. For Stage 1, we adopted the same control procedure as Exp-EmoPairwise (\textbf{Sec.\ref{sec:Exp-EmoPairwise}}). For Stage 2, two control trials (one per valence) required participants to select the correct concept set from three alternatives with an unambiguous ground-truth answer (\textbf{Fig.\ref{fig:S9}D}). 
For Stage 3, two additional control trials (one per valence) required participants to choose the correct concept set from two alternatives, one of which clearly corresponded to the target valence (\textbf{Fig.\ref{fig:S9}E}). Participants who failed more than one control trial in any stage were excluded from further analysis. After quality-control filtering, 47 participants were retained.

\noindent \textbf{Evaluation Criteria used in All Three Stages.}
Stage 1 evaluates the quality of generic affective rules using the same criteria as Exp-EmoInteract (\textbf{Sec.\ref{sec:Exp-EmoInteract}}), including validity, efficiency, preference, and fidelity. We report success rate, cumulative success rate, preference rate, SSIM-C, and L1-C (\textbf{Sec.\ref{sec:metric}}). Stage 2 evaluates the precision of personalized affective profiles by measuring preference rates for concept sets generated by different methods. Stage 3 evaluates the internal consistency of personalized affective profiles in EROS by measuring preference rates across concept sets constructed from accepted and rejected memories. Together, these three stages provide complementary evaluations of generic affective knowledge, personalized affective representations, and the stability of user-specific emotional profiles learned by EROS (\textbf{Sec.\ref{sec:res_emorule_stage1}--\ref{sec:res_emorule_stage23}}).

\subsubsection{Exp-EmoMotif: Cross-Scene Consistency in Motif Preference}
\label{sec:Exp-EmoMotif}

If the personalized motifs stored in EmoMem $\mathcal{H}$ capture genuine user preferences rather than image-specific biases, the same motif should elicit consistent responses when applied to visually and semantically related scenes. To evaluate this hypothesis, we conducted a human psychophysics experiment (Exp-EmoMotif) that measures the cross-scene consistency of motif preferences (\textbf{Fig.~\ref{fig:fig6}C1}).

\noindent \textbf{Experimental Protocol.} In each trial, participants were presented with one image from a semantically similar image pair together with two edited variants generated using two different motifs and a target valence. Participants selected the edited image that better evoked the target valence. At a later point in the experiment, the corresponding paired image was presented and participants repeated the same judgment task. To minimize ordering and contextual biases, paired images were not shown consecutively. Instead, evaluations from different image pairs were interleaved and trial order was fully randomized.

\noindent \textbf{Experimental Details.} Ideally, the personalized motifs stored in EmoMem $\mathcal{H}$ from Exp-EmoInteract (\textbf{Sec.\ref{sec:Exp-EmoInteract}}) could be reused directly in this experiment. However, doing so would substantially increase the duration of the in-lab study. We therefore designed Exp-EmoMotif as an independent experiment using EROS-1Loop and evaluated whether motif preferences generalized across semantically related scenes.

To construct semantically similar image pairs, we retrieved nearest-neighbor images from EmoSet using CLIP similarity. For each base image, two motifs were randomly sampled from $\mathcal{T}$ and used to generate two edited variants with EROS-1Loop. The same pair of motifs was then applied to the corresponding neighboring image, ensuring that preference consistency could be evaluated under identical affective intervention strategies.

For quality control, a minimum viewing time of 1 second was enforced for every trial. We computed the average response time for each participant and excluded participants whose mean response time was below this threshold. Although analyzed independently, Exp-EmoMotif was administered during the same in-lab session as Exp-EmoInteract (\textbf{Sec.\ref{sec:Exp-EmoInteract}}) to improve data-collection efficiency. During waiting periods in the interactive editing pipeline, participants completed interleaved Exp-EmoMotif trials. Participant quality was assessed separately for each experiment using their respective control procedures, and only participants who satisfied all quality-control criteria across both experiments were retained. The study involved the same 41 participants as Exp-EmoInteract. After quality-control filtering, data from 35 participants were retained, yielding a total of 6,214 trials.

\noindent \textbf{Evaluation Criteria.} We evaluate cross-scene consistency by measuring whether participants exhibit the same motif preference across semantically related image pairs. High consistency indicates that motif preferences generalize across scenes and reflect stable affective tendencies rather than image-specific responses. We quantify consistency using the alignment score (\textbf{Sec.\ref{sec:metric}}) and report the results in \textbf{Sec.\ref{sec:res_emomotif}}.

\subsection{Data Analysis}
\label{sec:data_analysis}
\subsubsection{Behavioral Metrics}
\label{sec:metric}
\noindent \textbf{Cumulative Success Rate.} For experiments involving iterative image editing, an edit is considered successful if the resulting image evokes the target valence according to human judgment. The cumulative success rate at loop $k$ is defined as the probability that a method achieves a successful edit within the first $k$ loops. This metric captures editing efficiency, where a higher cumulative success rate indicates that acceptable affective outcomes are achieved more rapidly.

\noindent \textbf{Preference Rate.} Given a set of $N$ candidate options, human observers select a single preferred option. The preference rate is defined as the proportion of trials in which a given option is selected. For example, in pairwise comparisons ($N=2$), the chance level is 50\%. A preference rate of 60\% for EROS indicates that it is preferred over the competing method in 60\% of trials. Higher preference rates therefore indicate stronger human preference for a given method or concept set.

\noindent \textbf{Alignment Score.} The alignment score measures agreement between two responses and is defined as the proportion of trials for which the responses are identical. The score equals 1 when the responses match and 0 otherwise. This metric can be used to quantify human--human agreement or human--machine agreement within the same task. Values range from 0 to 1, with higher scores indicating greater consistency.

\noindent \textbf{Intersection-over-Union (IoU).} Intersection-over-Union (IoU) measures the overlap between two image regions by dividing the area of their intersection by the area of their union:
$
\mathrm{IoU} = \frac{|A \cap B|}{|A \cup B|}.
$
IoU ranges from 0 (no overlap) to 1 (perfect overlap) and is used to quantify agreement between machine-predicted and human-annotated emotion-relevant regions.

\noindent \textbf{SSIM-C and L1-C.} Effective affective image editing should induce meaningful modifications within emotion-relevant regions while preserving the remaining image content. To quantify this trade-off, we introduce two contrastive metrics based on the Structural Similarity Index (SSIM)~\cite{wang2004SSIM} and the pixel-wise L1 distance. A schematic illustration of their computation is provided in \textbf{Fig.~\ref{fig:S7}}.

We first construct a spatial weight map $w \in [0,1]$ by aggregating binary masks from multiple human annotators and normalizing across pixels (\textbf{Sec.\ref{sec:Exp-EmoRegion}}). Regions with higher weights correspond to locations identified by humans as emotionally relevant.
For cross-valence editing, as illustrated in \textbf{Fig.~\ref{fig:S7}}, SSIM and L1 are computed separately over the preserved regions ($1-w$) and the emotion-relevant regions ($w$), yielding $SSIM_{\text{preserve}}$, $SSIM_{\text{edit}}$, $L1_{\text{preserve}}$, and $L1_{\text{edit}}$. In contrast, for same-valence editing, the emotion-relevant regions are expected to remain semantically unchanged while the surrounding context is enriched to strengthen the existing affect. Accordingly, the roles of $w$ and $1-w$ are reversed: the emotion-relevant regions ($w$) are treated as the preserved regions, whereas the remaining regions ($1-w$) are treated as the edited regions when computing the corresponding SSIM and L1 scores.
We then define
\[
\text{SSIM-C} = \frac{\text{SSIM}_{preserve}}{\text{SSIM}_{edit} + \varepsilon}, \qquad
\text{L1-C} = \frac{\text{L1}_{preserve}}{\text{L1}_{edit} + \varepsilon},
\]
where $\varepsilon$ is a small constant for numerical stability.
Higher SSIM-C indicates stronger structural preservation in the preserved regions relative to the edited regions. Lower L1-C indicates that pixel-level changes are concentrated within the intended editing regions while minimizing unintended changes elsewhere.

\subsubsection{Statistical Analyses}
\label{sec:stats_analysis}

\noindent \textbf{Bootstrapping and Distribution Analysis.} In each experiment, we assessed the statistical robustness of the results by generating empirical distributions of the model performances using bootstrapping. Specifically, we performed resampling with replacement for 5,000 iterations to obtain a stable distribution of bootstrap means for each condition. Welch’s two-tailed t-tests were then used to evaluate differences between conditions and to compare experimental results against chance levels, with statistical significance determined using a consistent threshold of $p < 0.05$. Throughout the manuscript, we used the term ``Welch’s two-tailed t-test" to assess statistical differences between two bootstrapped distributions.

\noindent \textbf{Significance levels.}
Results were considered statistically significant when $p < 0.05$, and non-significant (n.s.) otherwise. As numerical estimates of extremely small $p$-values can be unreliable, all values smaller than $10^{-3}$ are reported as $p < 10^{-3}$ (rather than exact values, e.g., $p = 10^{-40}$). This convention does not affect any of the conclusions.

\subsection*{Code and Data availability}
All the source code, AI models, and raw data are made publicly available in this submission through the following repository: \url{https://github.com/ZhangLab-DeepNeuroCogLab/EROS}

\section*{Competing interests}
The authors declare that they have no competing interests.

\section*{Financial Disclosure Statement}
This research is supported by the National Research Foundation, Singapore under its NRFF award NRF-NRFF15-2023-0001 and Mengmi Zhang's Startup Grant from Nanyang Technological University.

{\small
\printbibliography
}

\newpage
\section*{Main Figures}

\vspace{-6mm}

\newpage

\newpage

\newpage


\newpage

\newpage

\renewcommand{\thesection}{S\arabic{section}}
\renewcommand{\thefigure}{S\arabic{figure}}
\renewcommand{\thetable}{S\arabic{table}}
\setcounter{figure}{0}
\setcounter{section}{0}
\setcounter{table}{0}

\newpage
\section*{Supplementary Figures}
\begin{figure}[!ht]
\begin{center}
\includegraphics[width=\linewidth]{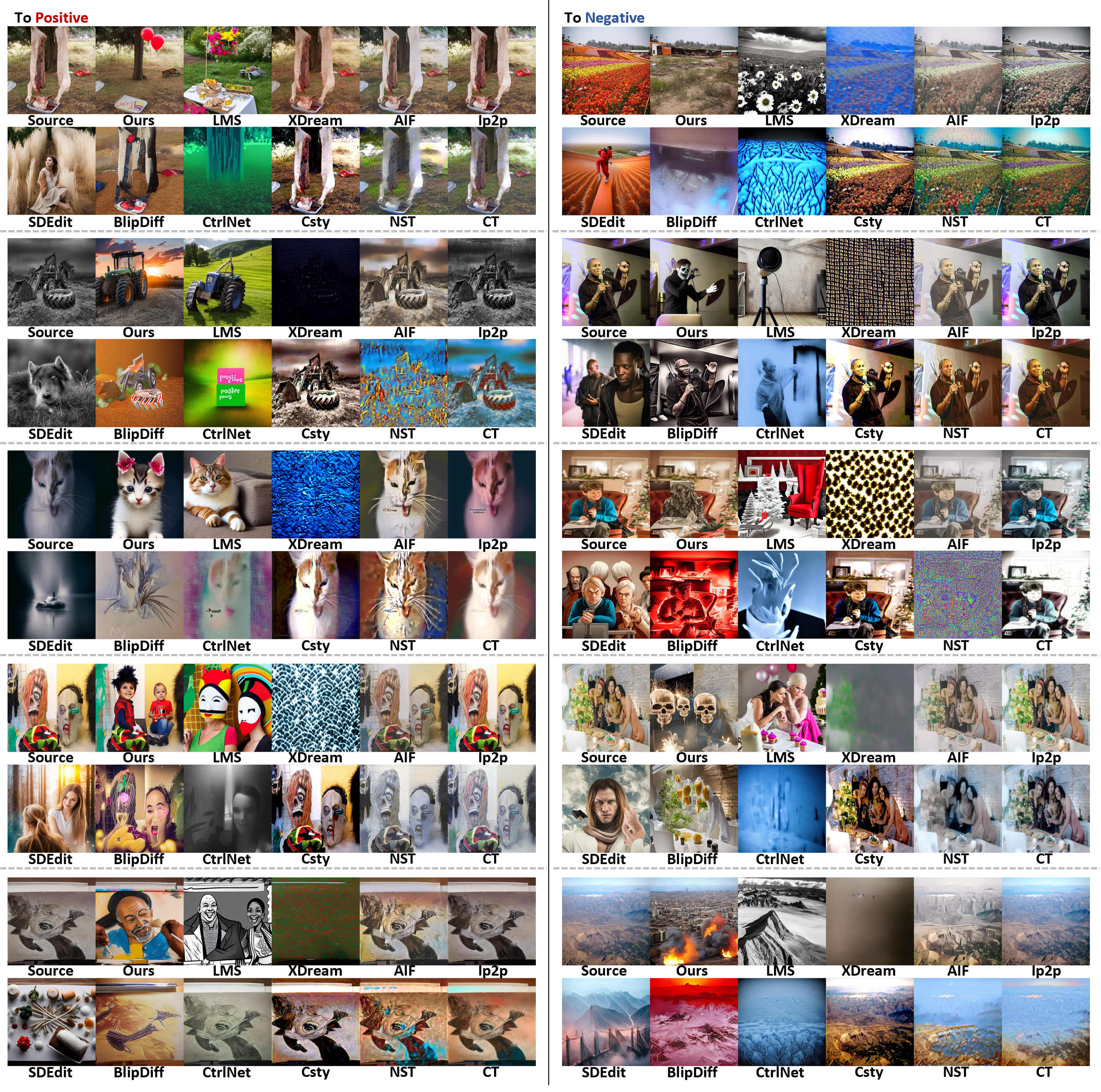}\vspace{-4mm}
\end{center}
   \caption{\footnotesize
    \textbf{Additional qualitative comparisons of affective image editing methods.} Representative editing results of EROS and all baseline methods for positive-valence (left) and negative-valence (right) editing tasks. Every two rows correspond to one source image and the corresponding edited images generated by EROS (Ours) and ten representative emotionally intelligent AI systems. The figure follows the same layout and visualization conventions as \textbf{Fig.~\ref{fig:fig1}F}. Zoom in for improved visualization. See \textbf{Sec.~\ref{sec:baselines}} for implementation details of the compared methods.
    }
    \vspace{-5mm}
\label{fig:S1}
\end{figure}

\newpage
\begin{figure}[!ht]
\begin{center}
\includegraphics[width=\linewidth]{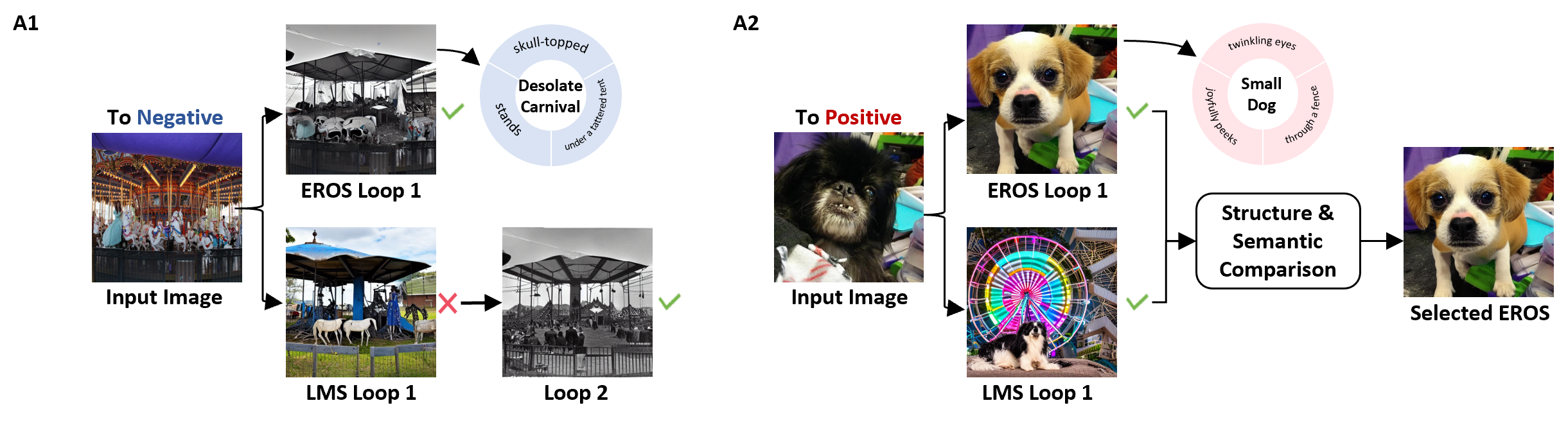}\vspace{-4mm}
\end{center}
   \caption{\footnotesize
    \textbf{Visualization of two example trials in the Exp-EmoInteract experiment (\textbf{Sec.~\ref{sec:Exp-EmoInteract}}).}
\textbf{A1,} Example in which EROS successfully evokes the target valence at the first editing loop (Loop 1), whereas LMS fails. The unsuccessful method (LMS) proceeds to the next editing loop to generate a new candidate image, successfully achieving the target valence in Loop 2. Green check marks and red crosses indicate whether participants judged the edited image at the corresponding loop to have successfully evoked the target valence. The arrow pointing to the blue motif in EROS denotes the affective editing rule retrieved from the EmoTree and applied to generate the edited image.
\textbf{A2,} Example in which both EROS and LMS successfully evoke the target valence at the first editing loop (Loop 1). The arrow pointing to the red motif in EROS denotes the affective editing rule retrieved from the EmoTree and applied during image editing. Participants subsequently perform a pairwise comparison and select the edited image that better preserves the semantic content and structural integrity of the source image while achieving the target valence. In this example, the image generated by EROS is selected as the preferred editing result.
    }
    \vspace{-5mm}
\label{fig:S3}
\end{figure}

\newpage
\begin{figure}[!ht]
\begin{center}
\includegraphics[width=\linewidth]{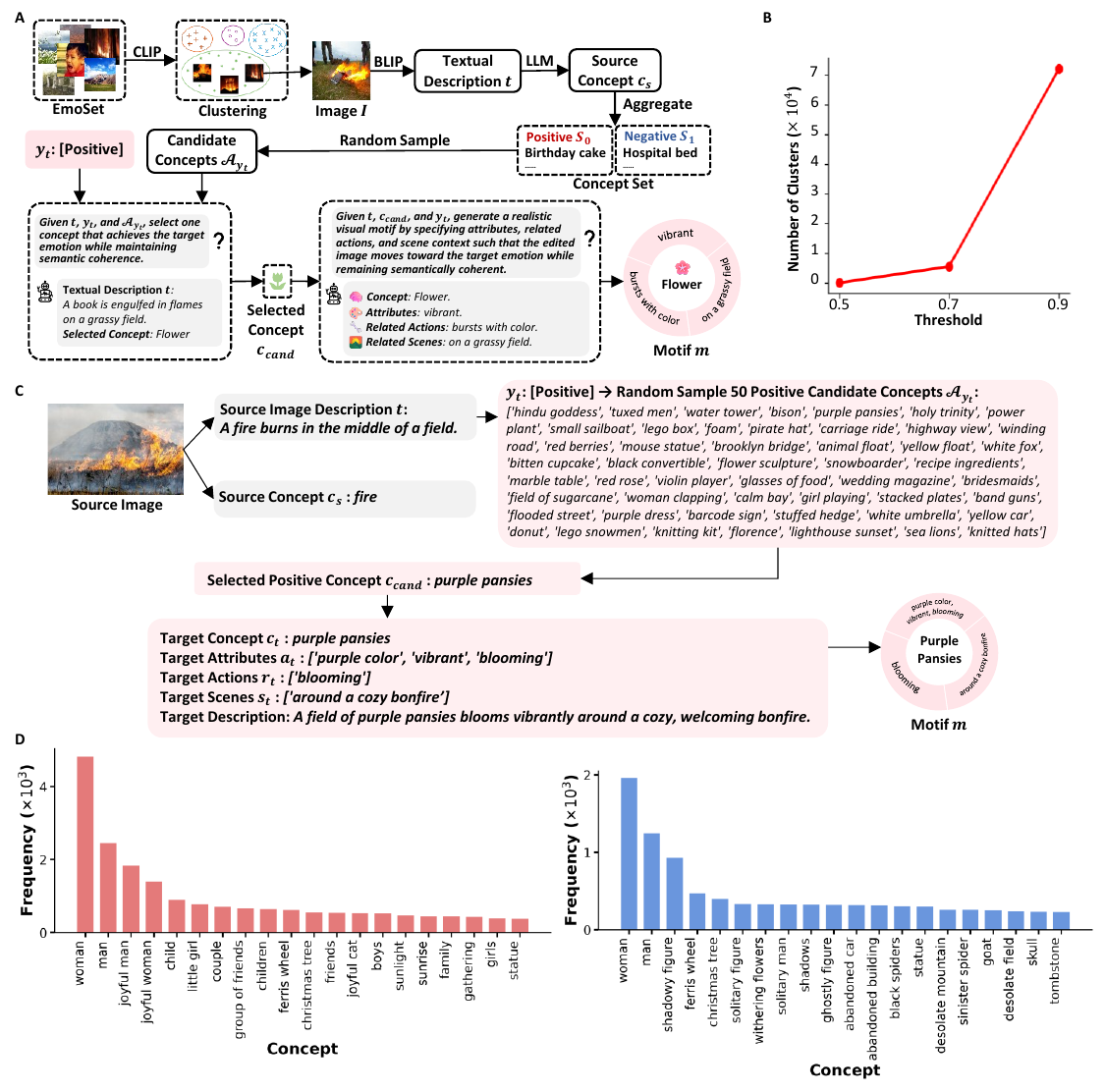}\vspace{-4mm}
\end{center}
   \caption{\footnotesize
    \textbf{Construction and analysis of the EmoTree $\mathcal{T}$ in EROS.}
    \textbf{A.} 
    Overview of the EmoTree construction pipeline. Images $I$ are organized into semantic clusters, from which source concepts $c_s$ and target concepts $c_t$ are extracted to generate compositional affective motifs $m$ for a given target valence $y_t$ consisting of concepts, attributes, actions, and scene contexts. Each motif defines an interpretable affective rule stored in the EmoTree. See \textbf{Sec.\ref{sec:module3_EmoTree}} for details.
    \textbf{B,} Effect of the cosine similarity threshold on semantic clustering in Level 1 of the EmoTree. Lower thresholds produce fewer, semantically broader clusters, whereas higher thresholds generate increasingly fragmented clusters. A threshold of 0.7 is adopted throughout EROS as a balance between semantic coherence and cluster diversity.
    \textbf{C,} Example of motif construction in Level 4 of EmoTree. Given a source image, BLIP first generates a textual description $t$, from which a large language model extracts the source concept $c_s$ (e.g., fire). Conditioned on a target valence $y_t$, EROS retrieves a semantically compatible target concept $c_{cand}$ (e.g., purple pansies) and constructs a compositional affective motif $m=(c_t,a_t,r_t,s_t)$ by generating associated attributes $a_t$, actions $r_t$, and scene contexts $s_t$. These components are combined to form an interpretable affective rule that guides emotion-aware image editing.
    \textbf{D.} The frequency distributions of the 20 most common target concepts aggregated across all affective motifs in the EmoTree. Positive-valence motifs (red) are dominated by socially positive and human-centered concepts, including woman, man, children, and group of friends. In contrast, negative-valence motifs (blue) span a broader range of concepts associated with peril, alienation, or degradation, including shadowy figure, withering flowers, abandoned building, and skull. This distribution indicates that positive affective rules converge on a relatively compact semantic space, whereas negative affective rules exhibit greater semantic diversity.
    See \textbf{Sec.\ref{sec:res_tree_prompt}} for more discussions.
    }
    \vspace{-5mm}
\label{fig:S4}
\end{figure}

\newpage
\begin{figure}[!ht]
\begin{center}
\includegraphics[width=\linewidth]{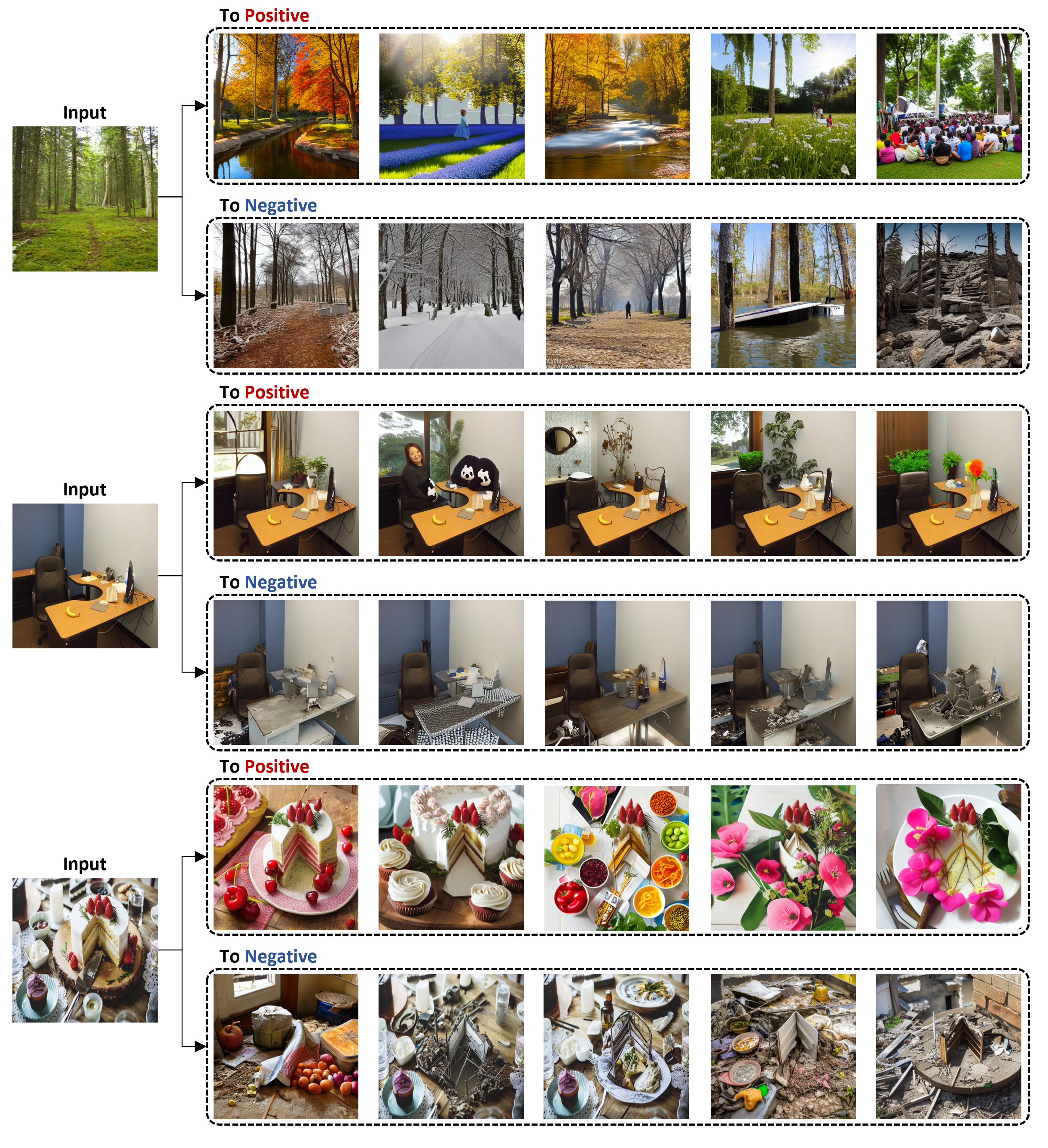}\vspace{-4mm}
\end{center}
   \caption{\footnotesize    
    \textbf{Examples of diverse affective image editing generated by EROS.}    
    Representative editing results produced by EROS for three input images under positive- and negative-valence targets. The leftmost column shows the input images. For each input, the upper row presents edits generated toward positive valence (\textit{To Positive}), whereas the lower row presents edits generated toward negative valence (\textit{To Negative}). Multiple edited outputs are shown for each target valence, illustrating the diversity of affective motifs retrieved from the EmoTree $\mathcal{T}$ while preserving semantic coherence with the source image. 
    The first two input images are sampled from the MSCOCO dataset~\cite{lin2014mscoco} and depict everyday naturalistic scenes. 
    The third input is sampled from the EmoSet dataset~\cite{yang2023emoset}.
    EROS supports both same-valence enhancement and cross-valence transformation. The emotion predictor $\mathcal{P}$ first estimates the valence of the input image (\textbf{Sec.~\ref{sec:module1_recog}}). During generation (\textbf{Sec.~\ref{sec:module4_gen}}), same-valence editing primarily enriches contextual regions while preserving the dominant emotion-relevant content of the source image. In contrast, cross-valence editing directly modifies emotion-relevant regions to shift the perceived emotional valence while maintaining the overall semantic content and structural integrity of the scene.
    }
    \vspace{-5mm}
\label{fig:S2}
\end{figure}

\newpage
\begin{figure}[!ht]
\begin{center}
\includegraphics[width=\linewidth]{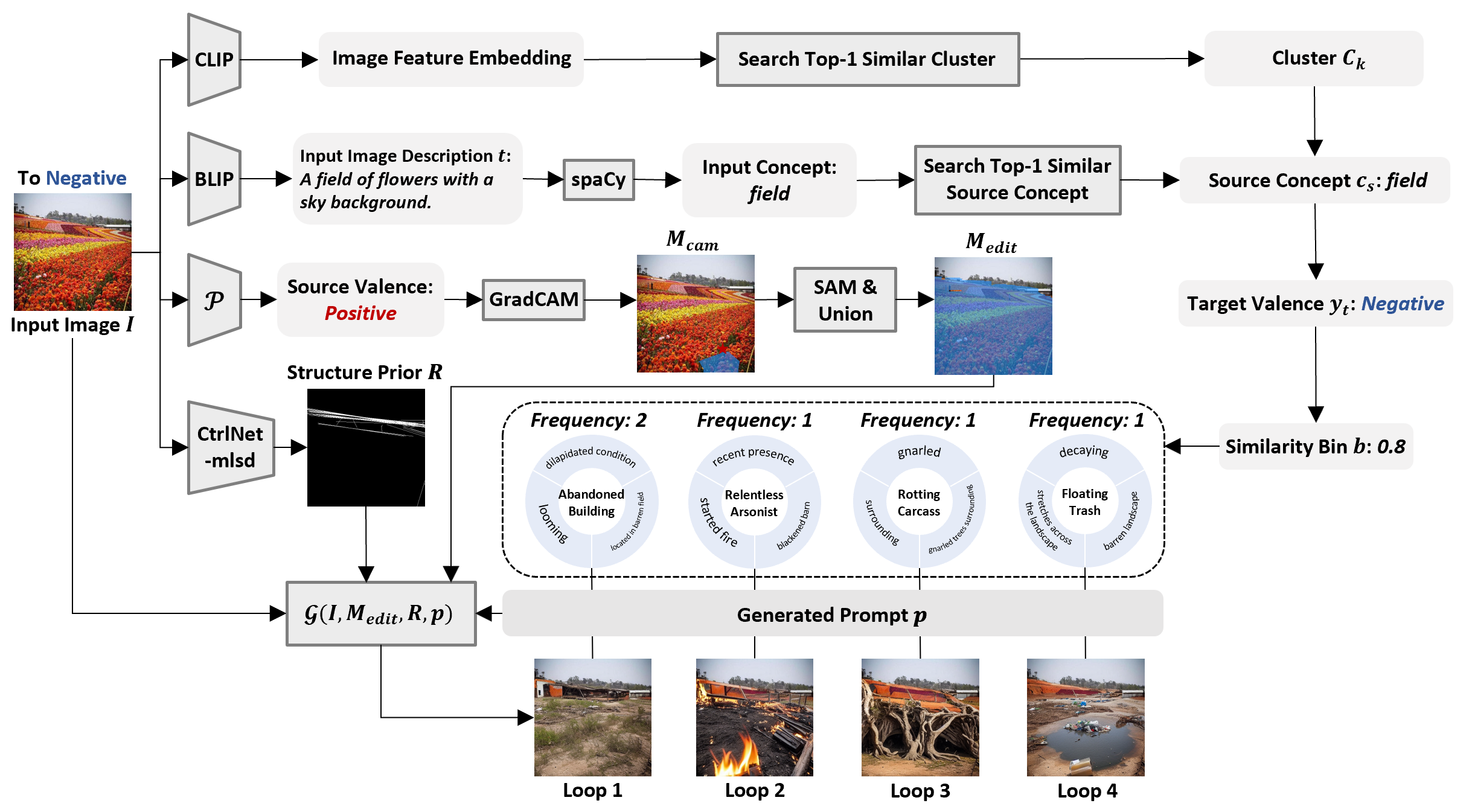}\vspace{-4mm}
\end{center}
   \caption{\footnotesize
\textbf{EROS inference pipeline for affective image editing without personalization.} Given an input image $I$ and a target valence $y_t$ (\textit{To Negative}), EROS first retrieves the most semantically similar cluster $C_k$ and source concept $c_s$ from the EmoTree using CLIP \cite{radford2021CLIP} image features, BLIP \cite{li2022blip} image captioning, and spaCy \cite{honnibal2020spacy} concept extraction. Candidate affective motifs are then retrieved from the corresponding target-valence branch and ranked according to semantic similarity and occurrence frequency (\textbf{Sec.~\ref{sec:module3_EmoTree}}). In parallel, the emotion predictor $\mathcal{P}$ estimates the source valence and localizes emotion-relevant regions, which are refined into an editing mask $M_{\text{edit}}$ for localized image manipulation (\textbf{Sec.~\ref{sec:module1_recog}}; \textbf{Sec.~\ref{sec:module2_region}}). Each retrieved motif is converted into a compositional text prompt and combined with the input image, editing mask, and structural prior $R$ to generate an edited image (\textbf{Sec.~\ref{sec:module4_gen}}). The bottom row illustrates successive editing loops. When an edited image is rejected, EROS retrieves an alternative affective motif and generates a new candidate. As personalization is disabled in this example inference process, motif selection is guided solely by the EmoTree without using the personalized memory bank EmoMem.   
    }
    \vspace{-5mm}
\label{fig:S5}
\end{figure}

\newpage
\begin{figure}[!ht]
\begin{center}
\includegraphics[width=\linewidth]{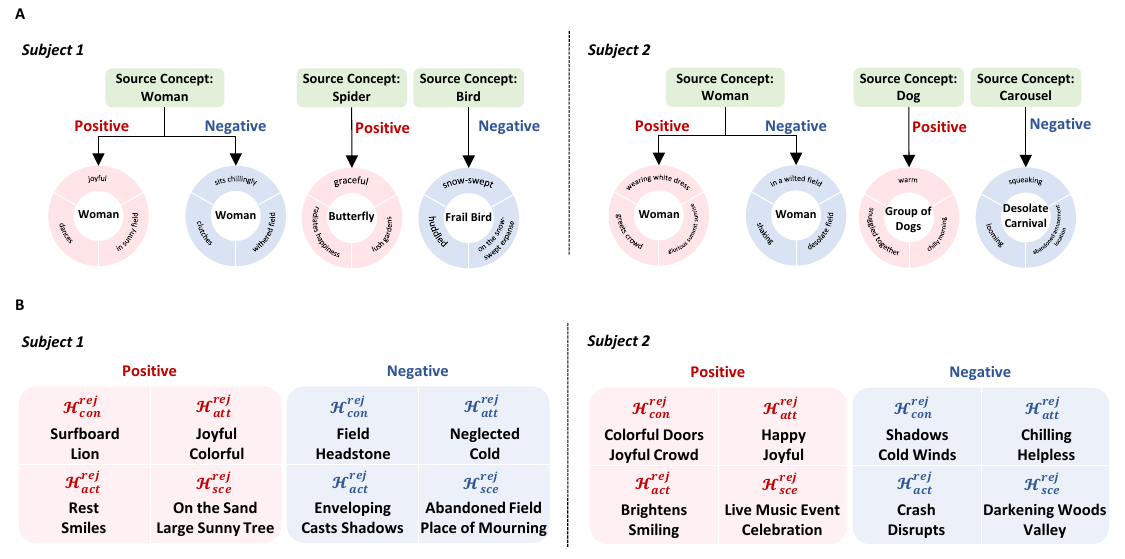}\vspace{-4mm}
\end{center}
   \caption{\footnotesize
   \textbf{Representative personalized affective profiles stored in EmoMem collected from individual participants in the Exp-EmoInteract experiment.} 
   \textbf{A.} Examples of accepted affective profiles stored in EmoMem from two participants in the Exp-EmoInteract experiment. For each participant, the figure shows the preferred motifs associated with different source concepts under positive- and negative-valence targets. While some source concepts are shared across participants (e.g., woman), the retrieved target concepts, attributes, actions, and scene contexts differ, illustrating that EmoMem captures user-specific preferences among multiple valid affective interventions.
   \textbf{B,} Examples of rejected affective profiles stored in EmoMem from two participants in the Exp-EmoInteract experiment. For each participant, the figure shows the rejected concepts, attributes, actions, and scene contexts associated with positive- and negative-valence conditions. 
   The differences across participants demonstrate that EmoMem captures user-specific dislikes.
    }
    \vspace{-5mm}
\label{fig:S6}
\end{figure}

\newpage
\begin{figure}[!ht]
\begin{center}
\includegraphics[width=0.5\linewidth]{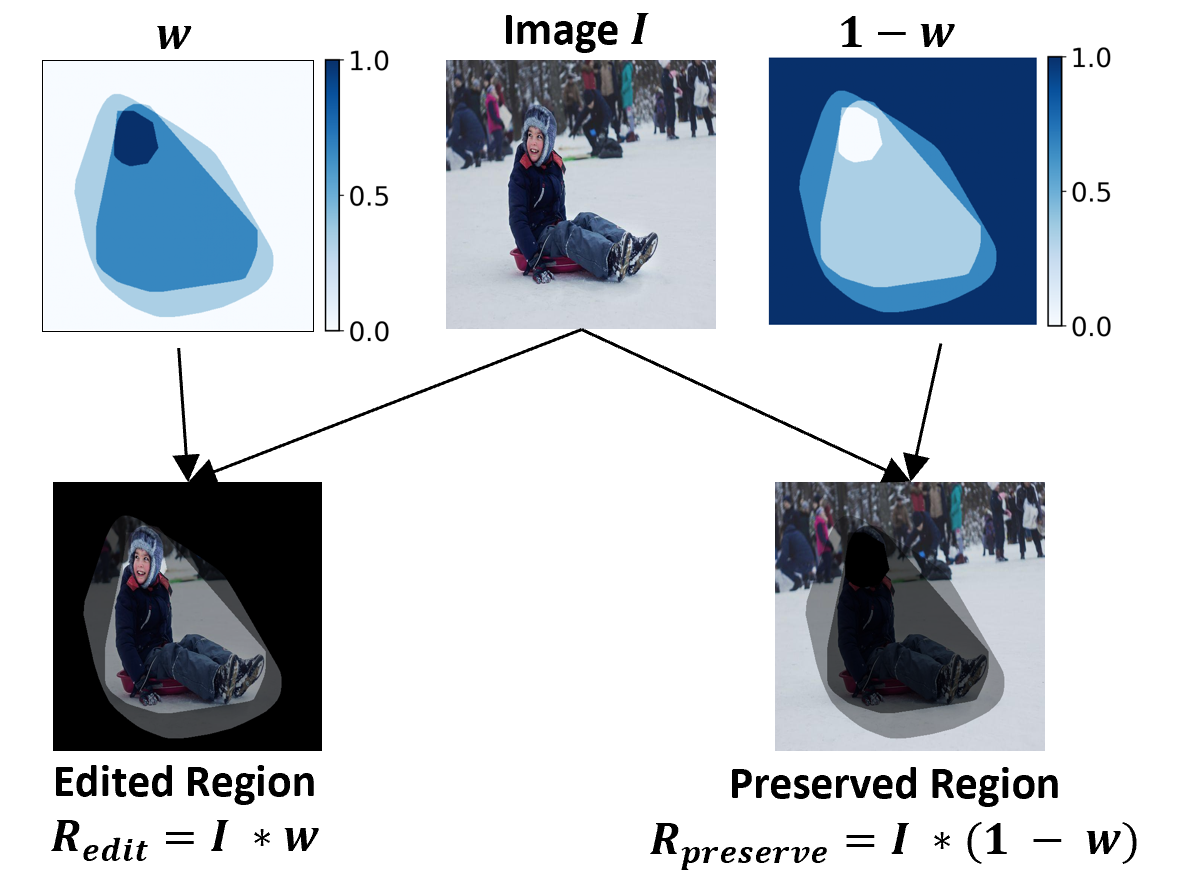}\vspace{-4mm}
\end{center}
   \caption{\footnotesize
   \textbf{Schematic illustration of the computation of SSIM-C and L1-C for cross-valence image editing.} Human annotations are aggregated into a normalized weight map $w \in [0,1]$, where larger values indicate emotion-relevant regions intended for modification and smaller values indicate regions expected to be preserved. The weight map and its complement $(1-w)$ are used to separately compute structural similarity (SSIM) and pixel-wise differences (L1) within edited and preserved regions, yielding the contrastive metrics SSIM-C and L1-C. See \textbf{Sec.\ref{sec:metric}} for the detailed introduction.
    }
    \vspace{-5mm}
\label{fig:S7}
\end{figure}

\newpage
\begin{figure}[!ht]
\begin{center}
\includegraphics[width=\linewidth]{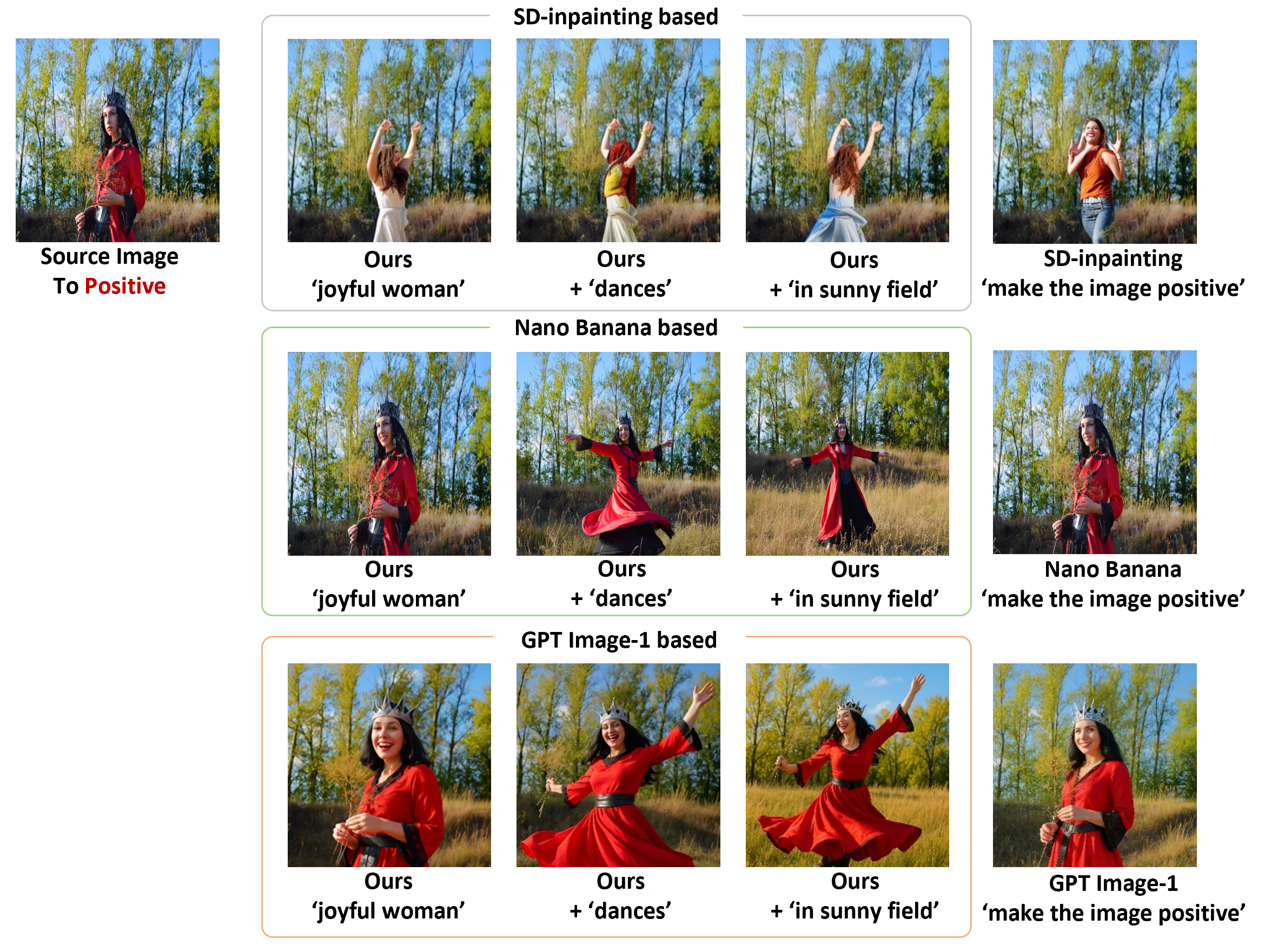}\vspace{-4mm}
\end{center}
   \caption{\footnotesize
\textbf{Generalization of EROS across different image generation models.} Representative editing results generated using three different image editing backbones with the same compositional text prompts produced by EROS: SD-Inpainting \cite{rombach2022SD} (top), Nano Banana \cite{nanobanana} (middle), and GPT Image-1 \cite{GPT-Image1} (bottom). For each backbone, EROS constructs compositional prompts by progressively incorporating different components of the retrieved affective motif, including the target concept and its attributes (joyful woman), the associated action (dances), and the scene context (in sunny field). The resulting edits are compared with a baseline prompt that directly specifies the target valence (``make the image positive"). Across all three image generation models, compositional prompts derived from the EmoTree produce more semantically coherent and emotionally expressive edits than directly prompting the desired emotion, demonstrating that EROS is model-agnostic and can be integrated with diverse text-conditioned image generation systems.   
    }
    \vspace{-5mm}
\label{fig:S8}
\end{figure}

\newpage
\begin{figure}[!ht]
\begin{center}
\includegraphics[width=\linewidth]{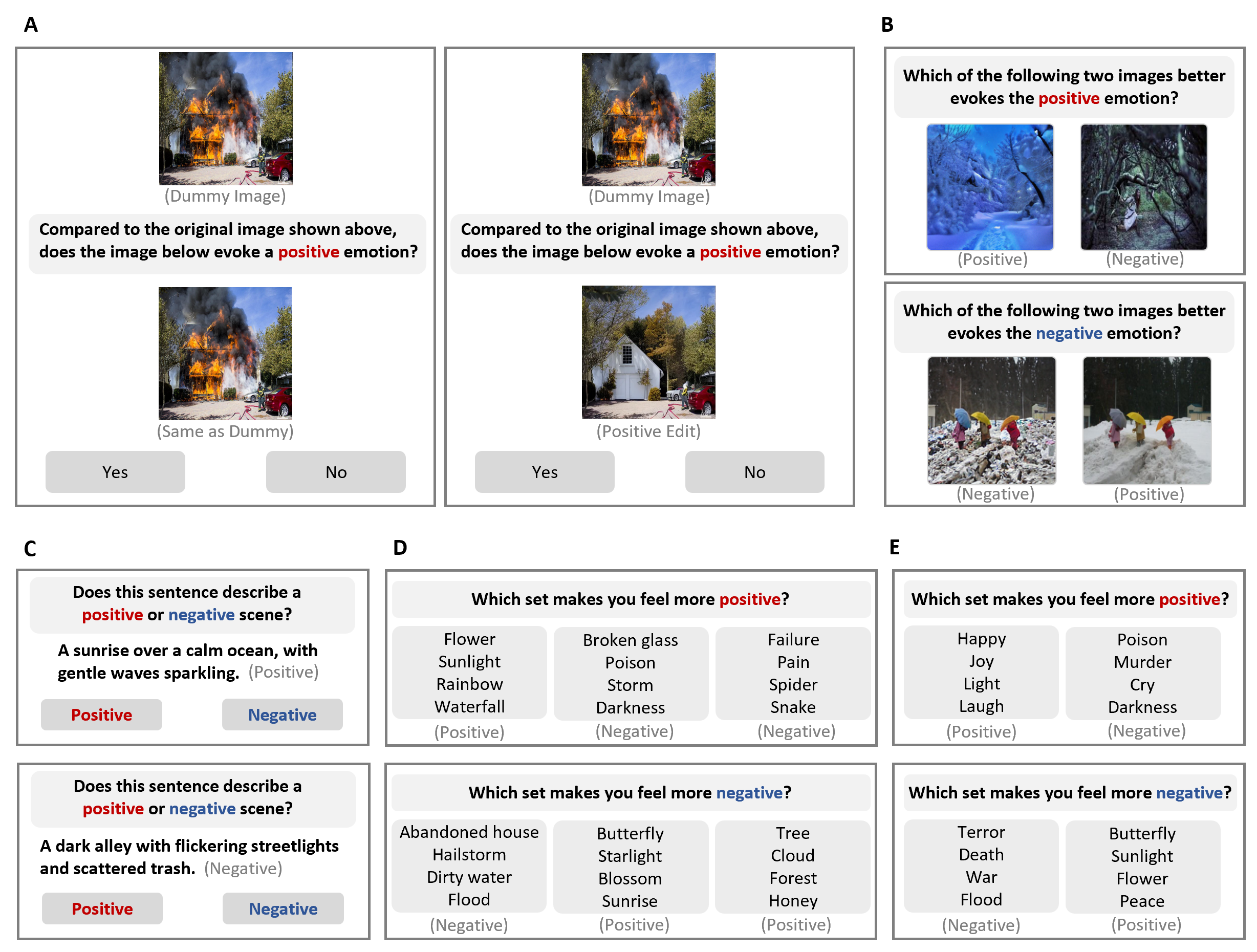}\vspace{-4mm}
\end{center}
   \caption{\footnotesize
    \textbf{Control trial schematics used across all the six human psychophysics expeirments for affective evaluation.}
    \textbf{A, Control trials for Exp-EmoInteract and Exp-EmoMotif.} 
    Participants determine whether an edited image evokes the target valence relative to the source image. One control image is identical to the source image (correct response: No), whereas the other is an unambiguous affective edit (correct response: Yes). 
    See \textbf{Sec.\ref{sec:Exp-EmoInteract}} and \textbf{\ref{sec:Exp-EmoMotif}} for details.
    \textbf{B, Control trials for Exp-EmoPairwise, Exp-EmoRegion and Exp-EmoRule Stage 1.}
    Participants select which of two images better evokes the specified positive or negative target valence. Each image pair consists of two images with clearly distinguishable and opposite emotional valences, providing an objective ground truth for quality control. Representative positive- and negative-valence control trials are shown in the upper and lower panels, respectively. See \textbf{Sec.\ref{sec:Exp-EmoPairwise}, \ref{sec:Exp-EmoRegion} and \ref{sec:Exp-EmoRule}} for details.
    \textbf{C, Control trials for Exp-EmoPrompt.}
    Participants classify whether a sentence describes a positive- or negative-valence scene. Each sentence conveys an unambiguous emotional valence, providing an objective ground truth for quality control. Representative positive- and negative-valence control trials are shown in the upper and lower panels, respectively.
    See \textbf{Sec.\ref{sec:Exp-EmoPrompt}} for details.
    \textbf{D, Control trials for Exp-EmoRule Stage 2.}
    Participants select the concept set that better evokes the specified target valence. Each trial presents three concept sets: one corresponding to the target valence and two corresponding to the opposite valence, providing an objective ground truth for quality control. Representative positive- and negative-valence control trials are shown in the upper and lower panels, respectively. 
    See \textbf{Sec.\ref{sec:Exp-EmoRule}} for details.
    \textbf{E, Control trials for Exp-EmoRule Stage 3.}
    Participants select the concept set that better evokes the specified target valence. Each trial presents two concept sets with clearly distinguishable and opposite affective valences, providing an objective ground truth for quality control. Representative positive- and negative-valence control trials are shown in the upper and lower panels, respectively.
    See \textbf{Sec.\ref{sec:Exp-EmoRule}} for details.
    For illustration only, the ground-truth labels are shown in gray brackets across all control trials (\textbf{A}--\textbf{E}); these labels were not visible to participants during the actual experiments.
    }
    \vspace{-5mm}
\label{fig:S9}
\end{figure}

\end{document}